\documentclass[10pt,journal,cspaper,compsoc]{IEEEtran}

\ifCLASSOPTIONcompsoc
  \usepackage[nocompress]{cite}
\else
  \usepackage{cite}
\fi

\ifCLASSINFOpdf
\usepackage[pdftex]{graphicx}
  \DeclareGraphicsExtensions{.pdf,.jpeg,.png,.tif}

\else
  \usepackage[dvips]{graphicx}
  \DeclareGraphicsExtensions{.eps}
\fi

\usepackage[cmex10]{amsmath}
\usepackage{amssymb}

\usepackage{array}
\usepackage{mdwmath}
\usepackage{mdwtab}
\usepackage{eqparbox}
\usepackage{longtable}
\usepackage{algorithm}
\usepackage{algorithmic}

\usepackage{multirow}

\usepackage[usenames,dvipsnames]{xcolor}

\newcommand{\comment}[1]{{}}
\newcommand{\todo}[1]{\textbf{\textcolor{purple}{[#1]}}}

\begin{document}

\title{On Detection of Faint Edges in Noisy Images}

\author{
Nati~Ofir, 
Meirav~Galun, 
Sharon~Alpert, 
Achi~Brandt, 
Boaz~Nadler, 
Ronen~Basri 
\IEEEcompsocitemizethanks{\IEEEcompsocthanksitem The authors are with the Department of Computer Science and Applied Mathematics, Weizmann Institute of Science, Rehovot, Israel.\protect\\
E-mail: see http://www.wisdom.weizmann.ac.il}
\thanks{}}


\IEEEtitleabstractindextext{%
\begin{abstract}
A fundamental question for edge detection in noisy images is how faint can an edge be and still be detected. In this paper we offer a formalism to study this question and subsequently introduce computationally efficient multiscale edge detection algorithms designed to detect faint edges in noisy images. In our formalism we view edge detection as a search in a discrete, though potentially large, set of feasible curves. First, we derive approximate expressions for the detection threshold as a function of curve length and the complexity of the search space. We then present two edge detection algorithms, one for  straight edges, and the second for curved ones. Both algorithms efficiently search for edges in a large set of candidates by hierarchically constructing difference filters that match the curves traced by the sought edges. We demonstrate the utility of our algorithms in both simulations and applications involving challenging real images. Finally, based on these principles, we develop an algorithm for fiber detection and enhancement. We exemplify its utility to reveal and enhance nerve axons in light microscopy images.
\end{abstract}

}

\maketitle

\IEEEdisplaynontitleabstractindextext

%
\IEEEpeerreviewmaketitle

\section{Introduction}

This paper addresses the problem of detecting faint edges in noisy images. Detecting edges is important since edges mark the boundaries of shapes and provide cues to their relief and surface markings. Noisy images are common in a variety of domains in which objects are captured under limited visibility. Examples include electron microscopy (EM) images (e.g., cryo-EM),
biomedical images with low tissue contrast, images acquired under poor lighting or short exposure time, etc. Low contrast edges also appear in natural images, where boundaries between objects may be weak due to shading effects, for example.

\begin{figure}[tb]
\begin{center}
\begin{tabular}{c c}
\fbox{\includegraphics[height=2.4cm]{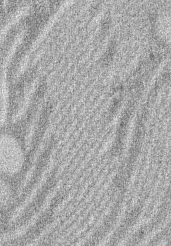}}~~
\fbox{\includegraphics[height=2.4cm]{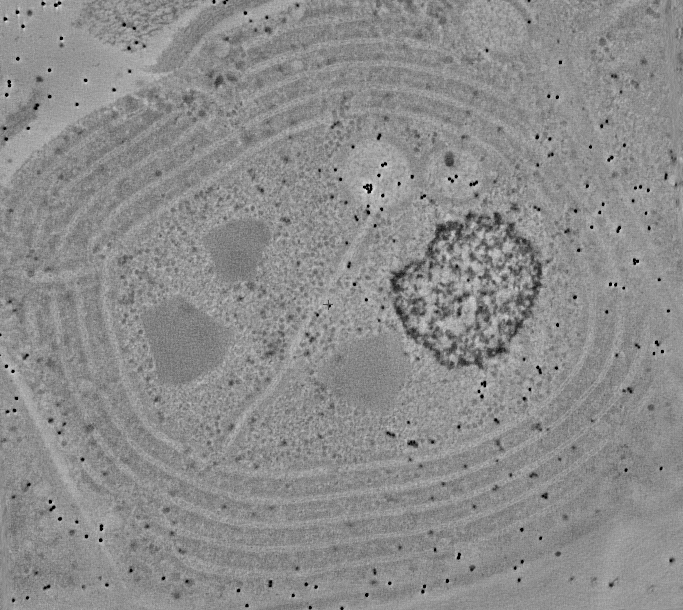}}\\
\end{tabular}
\end{center}
\caption{Electron microscopy images:
demonstrating the challenge of detecting edges embedded in noise.}
\label{fig:microscope}
%
\begin{center}
\begin{tabular}{r l}
\fbox{\includegraphics[height=2.4cm]{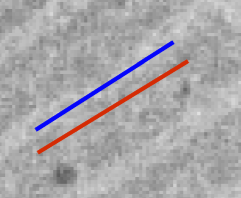}}~~
\fbox{\includegraphics[height=2.4cm]{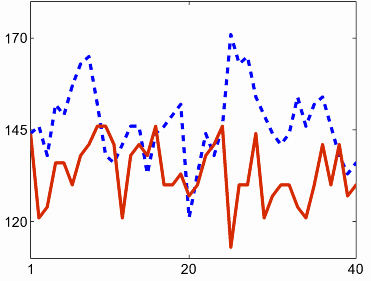}}\\
\end{tabular}
\caption{Two adjacent noisy intensity profiles (right) parallel to
a long edge (left) in an electron microscope image.
The noise leads to contrast reversals (locations where the red curve exceeds the blue one).} \label{fig:profiles}
\end{center}
\end{figure}

Noise poses a significant challenge because it leads to variability in the local contrast along an edge. Moreover, at low signal to noise ratio (SNR), defined as the edge contrast divided by the noise level, it may even lead to local \textit{contrast} \textit{reversals}. The images in Fig.~\ref{fig:microscope}, acquired by an electron microscope\footnote{We thank Eyal Shimoni and Ziv Reich for the  images.}, exemplify this kind of challenge. Despite the significant noise (see profile plots in Fig.~\ref{fig:profiles}), such low contrast edges are evident to the human eye due to their global consistent appearance over longer length scales.

As reviewed in Section~\ref{sec:previous-work}, most existing edge detection algorithms were designed to handle relatively low levels of noise. Common methods overcome noise by first smoothing the image, typically with a Gaussian filter. Such smoothing indeed reduces the noise, but may blur and weaken the contrast across edges, or even worse, blend adjacent edges. One approach to avoid these limitations is to apply a {\em matched filter}, that matches the curve traced by the sought edge. Such a filter separately averages the pixels values on each side of the edge to reduce the effect of noise, while maintaining the contrast across the edge. To apply these matched filters, however, the locations of the edges in the image need to be known, but this is precisely the problem of edge detection which we aim to solve.
Edge detection can thus be viewed as a \textit{search} for statistically significant filter responses in the space of feasible curves. In this paper we use this view to develop a statistical theory and novel computationally efficient algorithms for edge detection.

Our theory offers an answer to the following fundamental question: \textit{as a function of edge length and the complexity of the set of feasible curves, how faint can be the contrast of a step edge, and still  be detectable?} Specifically, given a family of feasible curves and their matched filters, we provide an approximate expression for the corresponding threshold that controls the false alarm rate of detecting spurious edges. We show that for certain relatively simple families of curves the detection threshold vanishes as the length of the curves tends to infinity. For such families, even the faintest edge can be detected provided it is sufficiently long. An important example is the set of all straight line segments, for which up to logarithmic factors,  the detection threshold decays polynomially  with the edge length \(L\) as $L^{-1/2}$.  In contrast, for certain exponential sets of curves we show that the edge detection limits are strictly positive.

Complementing our theoretical analysis, we introduce two new algorithms for edge detection. The first considers a family of straight lines at several lengths and orientations. Using a construction of Brandt and Dym~\cite{Dym99}, for an image of size $N=n^2$ pixels we present a computationally efficient $O(N \log N)$ algorithm that computes the corresponding matched filters and compares them with our theoretically derived threshold.
 The second algorithm considers a much larger (exponential) set of curved edges. We efficiently search this set of feasible curves using a multiscale binary-tree data structure, similar to that of Donoho and Huo\cite{Donoho02}. As this algorithm scans a considerably larger set of curves, its complexity is  $O(N^{1.5})$ operations in stringent version, and  $O(N\log N)$ in a greedy version.
 Finally, for better localization and removal of small misalignments, both algorithms conclude with a step of non-maximum suppression.

We demonstrate the detection performance of our algorithms on both simulated and real images. We also present its application to fiber detection and enhancement.
This paper presents in a unified manner and significantly extends earlier results that appeared in three conference publications \cite{Galun07}, ~\cite{Alpert10} and ~\cite{Ofir16}.

\section{Previous work}  \label{sec:previous-work}

Edge detection in images is a fundamental and well studied problem with many different algorithms and applications, see \cite{Bowyer,2011ivc_contour_survey} for detailed reviews.  Most methods, however, were designed to handle relatively low to medium levels of noise. Common edge detection algorithms overcome noise by first smoothing the image, typically using a Gaussian kernel.
A notable example is the Canny edge detector~\cite{Canny86}, which localizes edges by applying Sobel operators to the smoothed image to determine the gradient direction and follows this by hysteresis. Marr and Hildreth \cite{MarrHildreth} proposed a 2D Laplacian of a Gaussian filter to simultaneously smooth the image and detect zero crossings, see also \cite{kimmel2003regularized}. Scale-space representations extend this approach, and detect edges at multiple scales by combining spatially varying Gaussian smoothing with automatic scale selection~\cite{TonyLindeberg96,TabbAhuja1997}. A related approach was designed for fiber enhancement in medical applications~\cite{frangi98multiscale}. The main limitation of such isotropic smoothing is that it often reduces the contrast of weak edges, may blend adjacent edges and result in their poor localization.

To avoid smoothing across edges, while averaging the image intensity along them and thus enhance edge detection, powerful anisotropic diffusion (AD) and bilateral filter methods were developed, see
~\cite{PeronaMalik90, Weickert97,KimmelSochen2000} ,~\cite{TomasiManduchi98}
and references therein. While such methods considerably improve edge localization, their reliance on local gradients still limits their ability to detect edges in very noisy images.

Viewing AD as a denoising operation, an even more general approach is to first apply a sophisticated denoising algorithm to the input image, and then apply a simple local edge detector (such as Canny).
While seemingly appealing, unfortunately this approach is also not able to detect low contrast yet long edges. The reason is that current state-of-the-art denoising algorithms, such as non-local means~\cite{BuadesMorel2005} and medians~\cite{Singer} or  BM3D~\cite{Dabov}, are patch based with relatively small patch sizes, and due to the curse of dimensionality are not easily extended to larger patches \cite{LevinNadler}. Hence, as our experiments show, they are not suited for detecting long faint edges or dense fibers with small separation between them, of the type shown in Fig.~\ref{fig:microscope}.

\comment{
Of particular relevance to our work are the anisotropic diffusion (AD) methods. These methods utilize a diffusion tensor designed to avoid smoothing in the direction of the intensity gradients, while allowing smoothing in coherent directions. Indeed, while AD methods  considerably improve edge localization, their reliance on local gradients limits their ability to detect edges in very noisy images. Advanced AD methods (e.g., \cite{Weickert97,KimmelSochen2000}), therefore, suggest modifying the diffusion tensor through isotropic spatial averaging or resetting of its eigenvalues. However, such spatial smoothing and eigenvalue modifications are adapted to a single scale. Moreover, by using an averaged diffusion tensor, these methods accumulate squared \textit{local} intensity differences, and this may lead to smoothing across noisy, low contrast edges.
}

A different approach to image analysis is based on {\em curvelet} and {\em contourlet} image decompositions, which apply filters of various lengths, widths and orientations~\cite{Donoho2003,DoVetterli2005,Kalitzin99Invertible}. The main focus of these methods, however, is on  sparse representations of images and not on edge extraction or fiber enhancement. As discussed in a review of wavelet-based edge detection methods ~\cite{2012pami_wavelet_review}, wavelet transforms may
provide suboptimal representations of images with discontinuities embedded in low SNR.

Finally, to detect edges in natural images, ~\cite{RuzonTomasi2001} proposed to compute histograms of intensity, color and texture in two half disks on either sides of an edge. More recently, \cite{MartinFowlkesMalik2004,BEL2006} suggest supervised edge detection using features at multiple locations, orientations and scales. While these methods avoid smoothing across a candidate edge, the use of large scale features may lead to smoothing across nearby edges. As we demonstrate in Sec.~\ref{sec:experiments}, while these histograms are effective for detecting edges in natural images, they are far less so in noisy images. Furthermore, these histograms are calculated at all locations and orientations leading to slow algorithms. Related works (e.g.,~\cite{gPb2008,OWTUCM2009,gPbUCM2011,RenBo2012,MCG,Dollar}) extend the use of supervised learning for detecting natural images boundaries and combine information of global contours determined by an image segmentation process. A recent work ~\cite{Crisp} finds boundary edges in natural images by segmenting the image according to statistical dependencies between pixels.


On the theoretical side, a fundamental question is the study of limits on edge detection. In the statistical literature, such theoretical limits were studied for a model whereby an image is a discrete sampling on a 2D uniform grid of a function of two variables, which is piecewise constant (or more generally smooth), on two domains separated by a smooth boundary (the edge). In this setting, \cite{Tsybakov} derived minimax rates for edge detection and localization, in the limit as image size (number of pixels) tends to infinity, see also \cite{Donoho99}. A different line of work \cite{Hero_localization} considered the achievable accuracy in the localization of the edges.

As discussed in more detail below, the detection limit for real edges is closely related to the need to control the number of false detections. In this context,
~\cite{Morel2000,Morel2008} developed an a-contrario approach, stating that on a pure noise image an edge detection algorithm should output, with high probability, no edges at all. Following this principle, the authors proposed a method that for each candidate curve, counts the number of pixels with approximately consistent local gradient direction and finds the line segments as outliers in an otherwise random gradient model. A line segment detector (LSD) relying on this idea is presented in~\cite{LSD}. As discussed  in their paper, LSD is not well suited for images with very low SNR.

\section{Edge detection as search}

\subsection{Problem setup}


Let $I_0(x,y)$ denote a continuous noise-free function on a rectangular domain $\Omega \in \mathbb{R}^2$.
The observed image is a discrete and noisy version of \(I_{0}\), sampled on a rectangular grid
of size $N =n_{1} \times n_{2}$ pixels.  Assuming for simplicity ideal sampling with a $\delta$ point spread function,
the observed image pixels can be modeled as $I_{i,j} = I_0(x_i,y_j) + \xi_{i,j}$, where $\xi$ is additive noise.

Both in our theoretical analysis and in developing our algorithms, we make the following simplifying assumptions, though our algorithms work well even when these are slightly violated:
(i) the noise-free image \(I_0\) contains several step edges. Each step edge has a \textit{uniform} contrast (a constant gradient along the normal direction to it); (ii) adjacent edges in \(I_{0}\) are separated by a distance of at least \(w/2\) grid points; (iii) the
additive noise terms, \(\xi_{ij}\), are all i.i.d. zero mean Gaussian random variables with variance \(\sigma^{2}\).
Furthermore, we assume the noise level $\sigma$ is known. When unknown it can be estimated from the image ~\cite{Morel12}, \cite{Freeman2006}.

Given the noisy image \(I\), the problem of edge detection is to detect all the real edges in the original image \(I_0\), while controlling the number of detections of spurious edges.

\subsection{Search in a feasible set of curves }

\begin{figure}[t]
\begin{center}
\includegraphics[height=1.6cm]{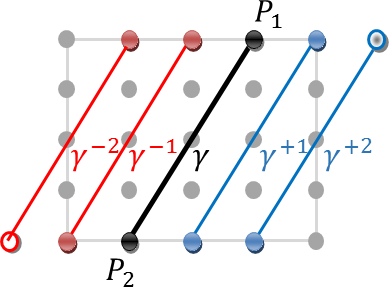}
\end{center}
\caption{Edge response for a straight curve: A straight curve $\gamma$ (in black) of length \(L=4\) connecting two grid points $P_1,P_2$. For a filter of width $w=4$,  its edge response is half of the difference between the average of \(wL/2\) interpolated pixel values on both sides of \(\gamma\) (denoted by $\gamma^{\pm i}$, $i \in {1,2}$. The corresponding filter is shown below in Figure~\ref{fig:Filters}.}
\label{fig:Straight_Response}
\end{figure}

As discussed in the introduction, for images with significant noise, it is very challenging, if not impossible, to detect edges based solely on local image gradients.
In our work we thus take a different approach: we assume that the edges in the clean image \(I_{0}\) belong to some large set of feasible curves \(\mathcal F\).
To each candidate curve \(\Gamma\in \mathcal F\) we associate an \textit{edge response}, \(R(\Gamma)\), defined as the value of the corresponding matched filter. Deferring precise definitions and implementation details to later sections, for a filter of width \(w\) applied to a curve of length \(L \) grid points, its edge response  \(R(\Gamma)\) is the difference of two averages of $wL/2$ interpolated pixel measurements on each side of the curve, see Fig. \ref{fig:Straight_Response} for an example.

Given the collection of edge responses \(\{R(\Gamma)\,|\,\Gamma\in\mathcal F\}\), we view the problem of edge detection as a {\em search} in this large collection of feasible curves, for sufficiently strong (e.g. statistically significant) matched filter responses.

\comment{
: Let \(f=(f_{1},\ldots,f_w)\) be a normalized filter mask of total width \(w\) (assumed fixed in our analysis), such that each \(f_{i}=\pm \tfrac1w\) and \(\sum_i f_i=0\). For example \(f=\tfrac{1}{{2}}(-1,1)\) for \(w=2\). We compute the filter response in the normal direction at each of the \(L\) pixels of the curve \(\Gamma_L\) and average them. That is, \(R(\Gamma_L)\) is proportional to the difference of two averages of $wL/2$ measurements on each side of the curve.
See fig ***.
}

\subsection{Outline of edge detection scheme}

Let ${\cal F}=\bigcup_{L=L_{\min}}^{L_{\max}}{\cal F}_L$ be the set of feasible curves, where  $\mathcal F_L$ is  the subset of candidate curves of length $L$, and $L_{\min},L_{\max}$ depend on the input image size.

This search perspective on the edge detection problem leads to the following generic framework. Given the noisy image \(I,\) the noise level \(\sigma\), the filter width \(w \) and the above set of feasible curves, edge detection proceeds as follows:

\begin{enumerate}
\item For each $L\in [L_{\min},L_{\max}]$
\begin{enumerate}
\item For each $\Gamma\in\mathcal F_L$, compute the corresponding edge response $R(\Gamma)$.
\item If $|R(\Gamma)| \le T=T( L,\mathcal F_L)$, discard $\Gamma$.
\item If $|R(\Gamma)|>T$, apply a consistent contrast test, to verify that $\Gamma$ indeed delineates an edge with a uniform contrast.
\end{enumerate}
\item Apply non-maximal suppression to the curves accepted in Step 1.
\end{enumerate}

The above general scheme leads to several interesting theoretical and practical questions. On the theoretical front, a key question is how should the threshold \(T\) be chosen, and consequently what is the faintest edge contrast detectable by this approach. On the practical side, the key challenge is to develop an algorithm to efficiently compute these potentially very large sets of edge responses.
In what follows we study these questions, and present two edge detection algorithms, one for straight edges
and one for curved edges.

\section{Detection threshold and minimal detectable contrast}  \label{sec:threshold}

We begin by addressing the first question above: \textit{Given the level of noise in the image, which curves can be discarded as not marking an edge?} Our aim is to derive a threshold that detects the faintest possible real edges, while controlling the false discovery of spurious edges having a large response only due to noise. Naturally, such a threshold should vary with the length of the considered curves, and comply with the following trade off. On the one hand we expect the threshold to decrease with curve length since by averaging along longer curves noise is more aggressively attenuated. On the other hand, if the number of curves in our considered collection grows with their lengths then so may the number of false detections; hence we may need to increase the threshold to control for this growth. This interplay between the length of a curve and the size of the search space determines the behavior of the threshold. In addition, our derivation can be used to infer the {\em minimal detectable contrast}, i.e., whether or not very faint edges can be detected {\em at all}.

\subsection{Detection threshold - derivation}

Given a collection of $K_L=|{\cal F}_L|$  candidate curves of length \(L\) and their edge responses, we wish to extract those that correspond to actual edges. As outlined above, we do this by keeping only those edge responses \(|R_{i}|>T\), where $T=T(L,K_L)$
is the \textit{detection threshold}, and afterwards applying additional tests and non-maximal suppression. Similar to the \textit{a-contrario} principle \cite{LSD},  and to multiple hypothesis testing in the statistical literature, a fundamental requirement for this detection threshold is that for a pure noise image \(I\) that contains no edges (e.g, \(I_0=const\)), the number of spurious false detections will be small.

To fix the rate of false detections we thus set $T$ such that for a pure noise image,
\begin{equation}  \label{eq:fix-fp}
P(R_{\max} \leq T ) \geq 1-\delta,
\end{equation}
where $R_{\max} = \max_{1 \le i \le K_L} |R_i|$ and $\delta$ is a small positive constant ($0<\delta\leq0.5$) independent of $L$.

Clearly, if a curve \(\Gamma_i\) passes through a constant intensity region of \(I_0\) away from its edges (e.g. a pure noise region of \(I\)), then its corresponding edge response \(R_{i}\) is a zero mean Gaussian random variable. If this curve is perfectly aligned with the cartesian grid, the \(wL\) measurements that enter the filter responses involve no interpolation and under our assumed image model are all i.i.d. \(\mathcal N(0,\sigma^2)\). In this case, \(R_i \sim \mathcal N(0,\sigma_L^2)$, where $\sigma_L^2=\sigma^2/wL$.

 For curves not perfectly aligned with the cartesian grid, the
\(wL\)  measurements are linear combinations of the original pixels and in general are not all statistically independent. Then, the corresponding variance slightly deviates from \(\sigma_L^2\) and depends on the precise shape of the curve. For simplicity, in our analysis we neglect this deviation.

Even with this simplification, the exact distribution of \(R_{\max}\)
is complicated and in general unknown. This is due to the complex dependencies of the edge responses of nearby or overlapping candidate edges.

To derive a threshold and gain insight regarding its dependence on the set of \(K_L\) candidate curves, we assume that all edge responses $R_i$ are independent and further that \(K_{L}\gg 1\) so that $T \gg \sigma_L$.
Under these assumptions
\begin{equation}  \label{eq:gproduct}
P(R_{\max} \leq T ) = \left[ P(|R_i| \leq T ) \right]^{K_L}.
\end{equation}
For $T \gg \sigma_L$,  we approximate the tail of the Gaussian distribution,
\begin{equation}
P(|R_i| \leq T ) \approx 1 - \sqrt{\frac{2}{\pi}}
\frac{\sigma_L}{T} \exp\left(-\frac{T^2}{2 \sigma_L^2}\right).
\end{equation}
Consequently, $P(R_{\max} \leq T ) \gtrapprox 1 - \delta$ implies
\begin{equation} \label{eq:approx}
\frac{\delta}{K_L} \gtrapprox \sqrt{\frac{2}{\pi}}
\frac{\sigma_L}{T} \exp\left(-\frac{T^2}{2 \sigma_L^2}\right).
\end{equation}
Taking the natural logarithm yields
\begin{equation}
\frac{T^2}{2 \sigma_L^2} \geq \ln\left(\frac{K_L}{\delta}\right) + \ln\left(\sqrt{\frac{2}{\pi}}\frac{\sigma_L}{T}\right).
\end{equation}
Ignoring the term $\ln(\sqrt{\frac{2}{\pi}} \frac{\sigma_L}{T})$ which is negative under the assumption that $T \gg \sigma_L$ and substituting for $\sigma_L$
we obtain the following approximate expression for the threshold
\begin{equation}  \label{eq:threshold}
T(L,K_L) \stackrel{\mathrm{def}}{=} \sigma \sqrt{\frac{2 \ln {(K_L/\delta)}}{wL}}.
\end{equation}

A key assumption in our derivation is that the filter responses of the $K_L$ feasible curves are statistically independent. In practice, this assumption does not hold, as curves may intersect or partly overlap. Therefore our threshold is conservative as the exact threshold may be lower. However, despite this simplification we show simulations on large sets of curves demonstrating a good fit to our predictions (see Fig.~\ref{fig:NoiseEstimation2}). As for the assumption that $T \gg \sigma_L$, note that in our expression~\eqref{eq:threshold} $T/\sigma_L=\sqrt{2 \ln {K_L/\delta}}$ which indeed is typically much larger than one.
For simplicity we consider a fixed $\delta=1/2$ and suppress its dependence.

Equipped with an expression for the threshold we proceed to study the detection limits of faint edges for different sets of curves. Of particular  interest is the asymptotic behavior of $T(L,K_L)$, when both the image size $N=n^2$ and the edge length $L$ approach infinity.   In general, for an image of size $N$  only curves of length $L \lessapprox N$ should be considered and in certain cases, e.g., when only straight lines are considered, only lengths $L \lessapprox \sqrt{N}$ are feasible.

Below we define two quantities that characterize the detectability of faint edges. The first is the {\em decay rate} of the threshold as a function of curve length,  captured by the ratio
\begin{equation}  \label{eq:ratio}
\rho_\alpha = \frac{T(L,K_L)}{T(\alpha L,K_{\alpha L})} \approx \sqrt{\frac{\alpha \ln (2K_L)}{\ln (2K_{\alpha L})}}
\end{equation}
with a constant $\alpha > 0$ typically set to $\alpha \in \{2,4\}$. The second quantity is the limiting value of the threshold,
$$
T^\infty = \lim_{L,N \rightarrow \infty} T(L,K_{L}).
$$

We distinguish between two cases. When $T^\infty > 0$ edges with contrast lower than $T^\infty$ cannot be detected reliably. In this case longer filters do not improve detection. This is the case for the family of all possible curves, as we prove in the next section. Conversely, when $T^\infty = 0$, the threshold typically decays at a polynomial rate, $T(L,K_L)=O(1 / L^{\log_\alpha \rho_\alpha})$. In this case in theory even the faintest edge can be detected, provided it is sufficiently long. Later on  we provide examples of families of curves (e.g., straight line segments) for which $T^\infty$ vanishes.

\subsection{Lower bound for the full set of curves}
\label{sec:full-search}

A basic question is whether very faint edges can be detected if we consider the full set of general, non-self intersecting discrete curves. Obviously this set is exponential in $L$, since from any starting pixel, the number of upper-right monotone curves in a 4-connected lattice is  $ 2^L$, and monotone curves form a subset of the non-self intersecting curves. While our analysis above implies that $T^\infty$ does not vanish for exponential sets of curves, it is based on the assumption that different responses are statistically independent.

The following argument shows that indeed when the full set of curves is considered, $T^\infty$ is strictly positive. We prove this by deriving a lower bound on $T^\infty$ for the subset of monotone curves. A curve is monotone if its tangent vectors at all points along the curve lie within one quadrant. For simplicity, we show this on an 8-connected lattice and with the particular $3 \times 1$ filter of width \(w=2\), $f=\tfrac12(1,0,-1)^T$. The result can be extended to lattices with different number of connections and to wider filters of different orientations.

To this end, let $I$ be a pure noise image, and let $p_0$ be any pixel
at distance at least $L$ from the boundary of $I$. We then have the following result:

{\bf Lemma:} {\em There exists a monotone curve $\Gamma=\Gamma(I)$ of length $L$ starting
at $p_0$, such that
\begin{equation}
\mathbb{E}_I[R(\Gamma)]=\frac{\sigma}{\sqrt{2\pi}}>0
        \label{eq:Lower_Bound}
\end{equation}
and such that its variance is $O(1/L)$. }

{\em Proof:} Consider the following greedy approach to select a monotone curve with  high response. Starting at  the initial pixel $p_0$, at each grid point of the curve we consider moving one step either to the right ("east") or upper-right ("north east") and select the direction that yields a higher response. Specifically, assuming the current point on the curve \(\Gamma \) is $p_i=(x_i,y_i)$, we select the next grid point $p_{i+1}$ by comparing the responses at $(x_i+1,y_i)$ and $(x_i+1,y_i+1)$ and choosing the one with larger response. By definition, the value of the filter applied at these two points is
\begin{eqnarray}
r(x_i+1,y_i)&\!=\!&\tfrac12(I(x_i+1,y_i+1)-I(x_i+1,y_i-1)) \nonumber\\
r(x_i\!+\!1,y_i\!+\!1)&\!=\!&\tfrac12(I(x_i+1,y_i+2)-I(x_i+1,y_i)). \nonumber
\end{eqnarray}
Therefore these two responses are independent Gaussian variables with zero mean and variance $\sigma^2/2$. Their maximum, denoted $r_{i+1}$, is a random variable with mean $\sigma/\sqrt{2\pi}$ and variance $\frac{\sigma^2}2(1-1/\pi)$~\cite{Nadarajah08}.

Let $\Gamma$ denote the curve passing through the $L$ points $p_1,p_2,...,p_L$ selected by this process. By definition, its edge response is the average of its $L$ edge filters. Hence,
\[\mathbb{E}_I[R(\Gamma)] = \frac1L\sum_{i=1}^L \mathbb{E}_I[r_i]=\frac{\sigma}{\sqrt{2\pi}}.\]
Since by construction, the responses $r_i$ are independent,
\[\mathrm{Var}_I[R(\Gamma)] = \frac1{L^2}\sum_{i=1}^L \mathrm{Var}_I[r_i] = \frac{\sigma^2}{2L}\left(1-\frac1\pi\right).\]
\hfill$\Box$

This lemma implies that as $L\to\infty$, $T^\infty$ is indeed strictly positive. The reason is that using the strong law of large numbers, as $L \rightarrow \infty$ the probability that the response for $\Gamma$ will obtain the value $\sigma/\sqrt{2\pi}$ approaches 1. This shows that for any reasonable value of $\delta$ (the false detection rate) $T^{\infty}$ must be strictly positive (and $\ge \sigma/\sqrt{2\pi}$). Consequently, faint edges with contrast lower than $T^{\infty}$ cannot be detected unless we accept a considerable number of false positives. Finally, note that the main arguments in this section extend also to other (non-gaussian) i.i.d.\ noise models.


\section{Consistent contrast test}
\label{sec:consistent_contrast_test}
The threshold $T(L,K_L)$ in \eqref{eq:threshold} is designed to reject curves whose responses are produced purely by noise. If the response of a candidate curve $\Gamma$ of length $L$ satisfies $|R(\Gamma)| < T(L,K_L)$ then we conclude that $\Gamma$ travels completely ``between edges.'' If however $|R(\Gamma)| \ge T(L,K_L)$ we conclude that, with high probability, $\Gamma$ traces an edge {\em at least part of its way}. In particular $R(\Gamma)$ can be the result of averaging a very high contrast edge on a small part of   $\Gamma$ with a zero contrast over the remaining part of the curve.  Our next task, therefore, is to identify those curves that depict edges throughout their whole extent and distinguish them from responses due to fragmented edges. We achieve this by applying a {\em consistent contrast test}, as we explain below.

In designing the consistent contrast test we assume that a true edge has approximately a constant contrast. We view the response $R(\Gamma)$ of a curve $\Gamma$ of length $L$ and width $w$, as an average of $L$ length-1 responses $r_1,...,r_L$. We model the distribution of each local response $r_i$ as follows:
\begin{enumerate}
\item If point $i\in\Gamma$ depicts a non-edge (e.g., a gap) then $r_i\sim N(0,\sigma^2/w)$.
\item If point $i\in\Gamma$ belongs to a true edge fragment then the expectation of  $r_i$ equals the unknown contrast of the edge. We then assume that $r_i$ is normal with unknown mean $\mu_e$ and standard deviation $\sigma_e/ \sqrt{w} $. We further assume that $\sigma_e \ge \sigma$. Note that $\sigma_e=\sigma$ implies that the edge contrast is exactly constant.
\end{enumerate}
The consistent contrast test distinguishes between two hypotheses: whether $\Gamma$ depicts an edge in its entire extent, in which case we expect $r_i$ for all $i$ to be drawn from the edge distribution, or $\Gamma$ contains some non-edge portions, in which case some of the $r_i$'s follow the noise distribution.

To apply the test we first estimate the unknown edge mean $\mu_e$ and standard deviation $\sigma_e$ from the pixels surrounding the edge. We then divide $\Gamma$ into short segments of length $\ell$ for some constant $\ell \ge 1$. For each segment $S$ we test if it is more likely to come from the edge distribution or from noise, i.e., we test whether $P(R(S)|\mathrm{edge}) > P(R(S)|\mathrm{noise})$ and we accept $\Gamma$ only if {\em all} of its segments are more likely as edges than noise. Taking the length of $S$ into account, $P(R(S)|\mathrm{edge})$ is Gaussian with mean $\mu_e$ and standard deviation $\sigma_e/\sqrt{w\ell}$ and $P(R(S)|\mathrm{noise})$ is Gaussian with zero mean and standard deviation $\sigma/\sqrt{w\ell}$. Assuming the prior $P(\mathrm{edge})=P(\mathrm{noise})=0.5$, the corresponding threshold \(b(\Gamma)\) of the generalized likelihood test  satisfies
\begin{equation}  \label{eq:consistent}
(\sigma_e^2 - \sigma^2)b^2(\Gamma) + 2 \sigma^2 \mu_e b(\Gamma) + \frac{2}{wl} \sigma^2 \sigma_e^2 \ln (\frac{\sigma}{\sigma_e}) -\sigma^2 \mu_e^2 \geq 0
\end{equation}
In general this quadratic equation gives two thresholds of opposite signs. In our implementation we only use the positive one. If  $\sigma_e = \sigma$, the optimal threshold is $b(\Gamma) = \mu_e/2$.


\section{Detection of straight edges}  \label{sec:straight}

We now introduce two algorithmic realizations of our statistical theory and design computationally efficient edge detection algorithms for straight and curved edges, respectively.


We begin by defining our family of straight line segments and then explain how their corresponding matched filter responses can be efficiently computed in a hierarchical way.
It should be emphasized that   the detection threshold approaches zero as the length of the lines increases.

\subsection{Filter definition}

For simplicity, we  consider only line segments whose endpoints are integral.
In principle, the family of all straight lines
 connecting any two pixels in the image, is of size \(O(N^2)\), which can be too large and slow to compute all of their matched filters. Instead,  using a construction by Brandt and Dym~\cite{Dym99} we consider only a subset of them of size  $O(N \log N).$ As we describe below, this allows for fast computation, with little loss in accuracy.

We next define the family of straight line filters. Each filter in our family  takes the shape of a parallelogram with (at least) one axial pair of sides. Consider a line segment beginning at  pixel $(x_i,y_i)$ and terminating at $(x_j,y_j)$, and denote the angle that the segment forms with the $X$-axis by $\theta$ ($-\pi/2 < \theta \le \pi/2$). We use the max norm to define its length $L=\max(|x_j-x_i|,|y_j-y_i|)$. We  distinguish between two sets of lines according to their orientations; {\em near-vertical lines} are those for which $|\theta| > \pi/4$, and {\em near-horizontal lines} are those for which $|\theta| < \pi/4$. Below we shorthand these to {\em vertical} and {\em horizontal} lines. The case of $|\theta|=\pi/4$ can either be assigned arbitrarily, or, as in our implementation, to {\em both} sets (in which case two {\em different} types of filters will be used for candidate edges at these specific  orientations).

Below we consider the family of filters of vertical orientations. The horizontal filters are defined analogously as $\pi/2$-rotations of the vertical filters.  Our  vertical filter computes half the difference between the mean intensities of two congruent parallelograms with a pair of horizontal sides. The height of these parallelograms is denoted by $L$; the horizontal sides are each of length $w/2$, and the other pair of sides form an angle $\theta$ with the $X$-axis. Formally, let
\begin{eqnarray}
&& M(x_0,y_0,L,w,\theta) = \label{eq:OrientedMeans} \\
&& ~~~~~ \frac{1}{wL}\int_{0}^{L} \int_{x_0+y/\tan\theta}^{x_0+y/\tan\theta+w} I(x,y) dx dy. \nonumber
\end{eqnarray}
Then, a straight vertical filter is defined by
\begin{eqnarray}
&&F(x_1,y_1,L,w,s,\theta) = \frac{1}{2}M(x_1+\frac{s}{2},y_1,L,\frac{w}{2},\theta)
\nonumber\\
&&- \frac{1}{2}M(x_1-\frac{s}{2}-\frac{w}{2},y_1,L,\frac{w}{2},\theta).
\end{eqnarray}
Throughout the paper we assume that $L$, $w/2$ and $s$ are positive integers, and that the latter two are constant.


\begin{figure}
\centering
\includegraphics[width=2.3cm,height=1.5cm]{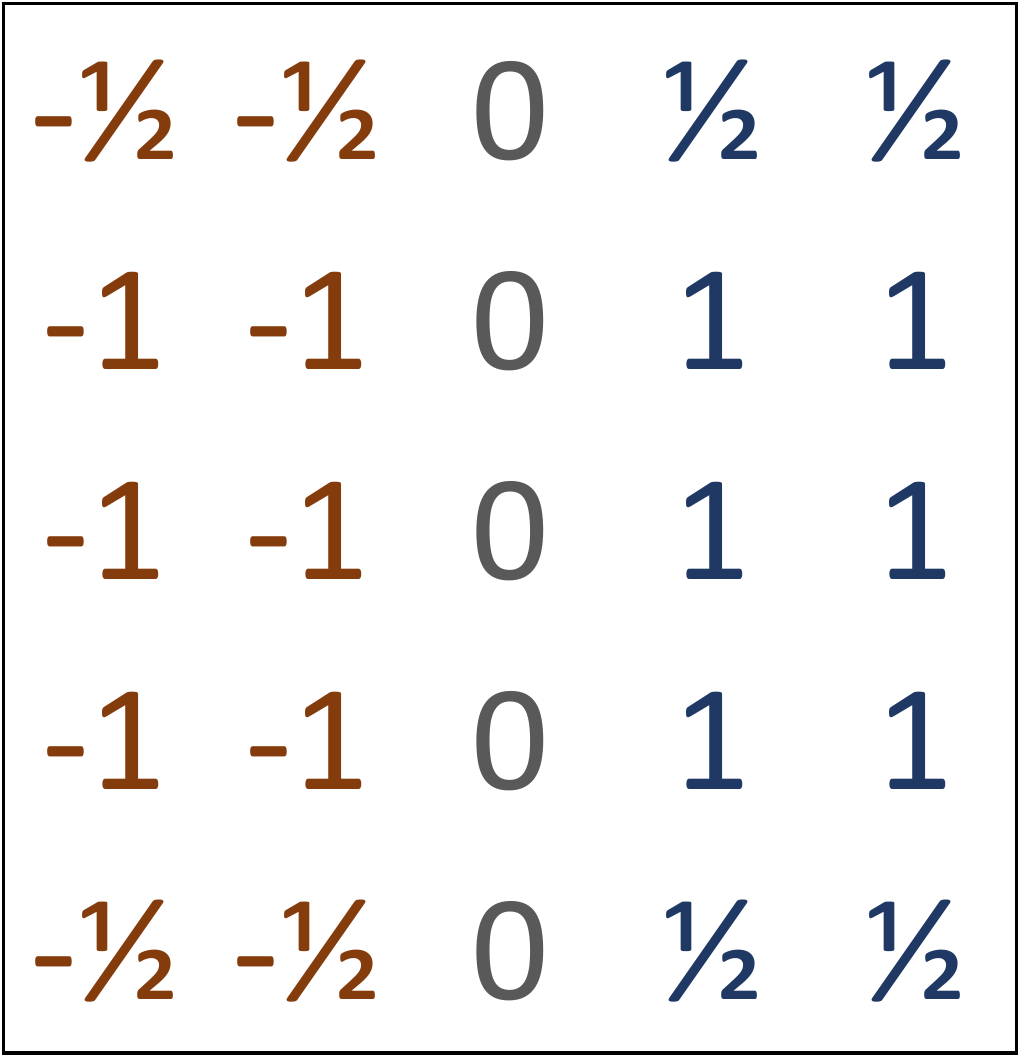}
\includegraphics[width=2.9cm,height=1.5cm]{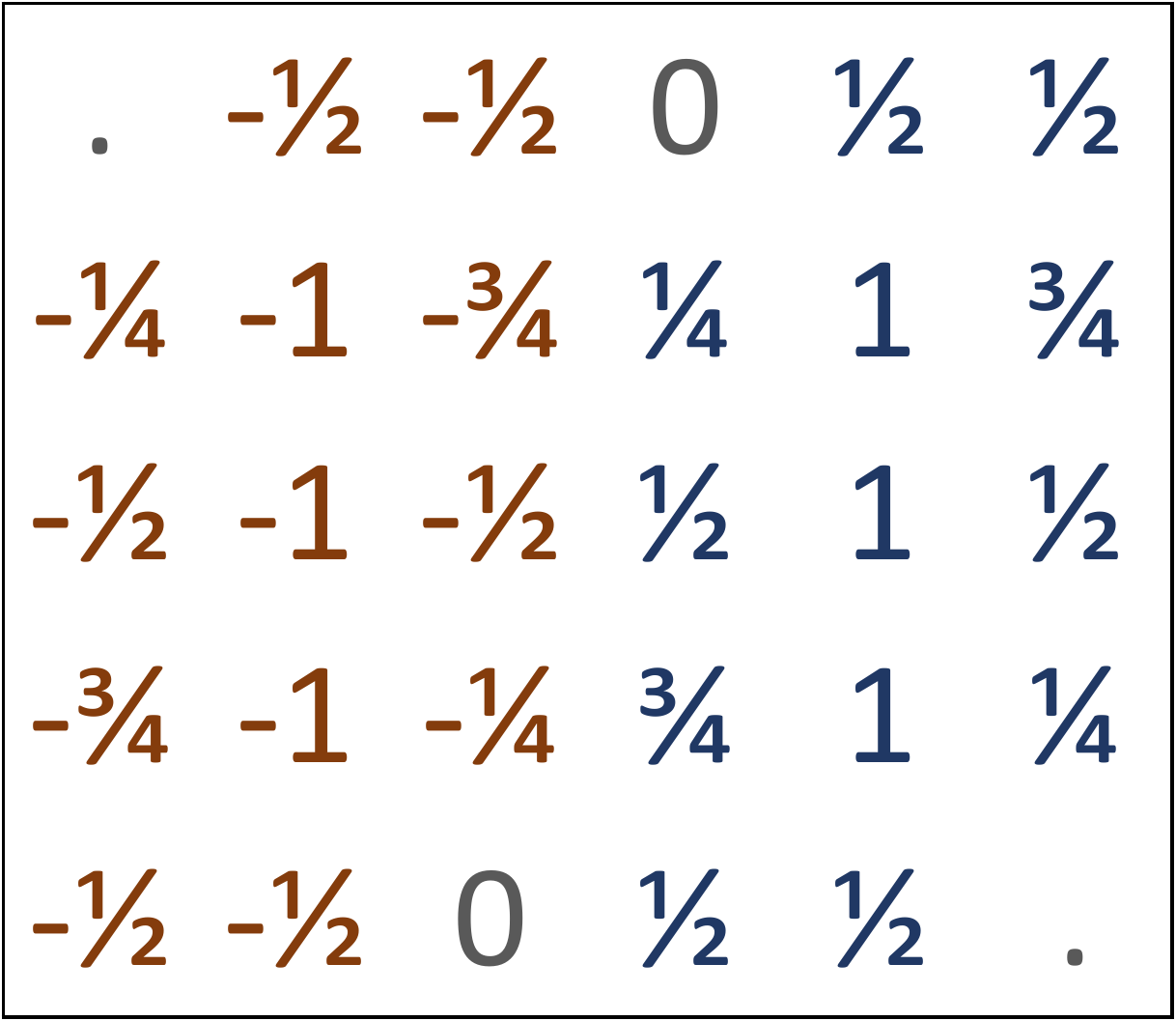} \\
\includegraphics[width=3.0cm,height=1.5cm]{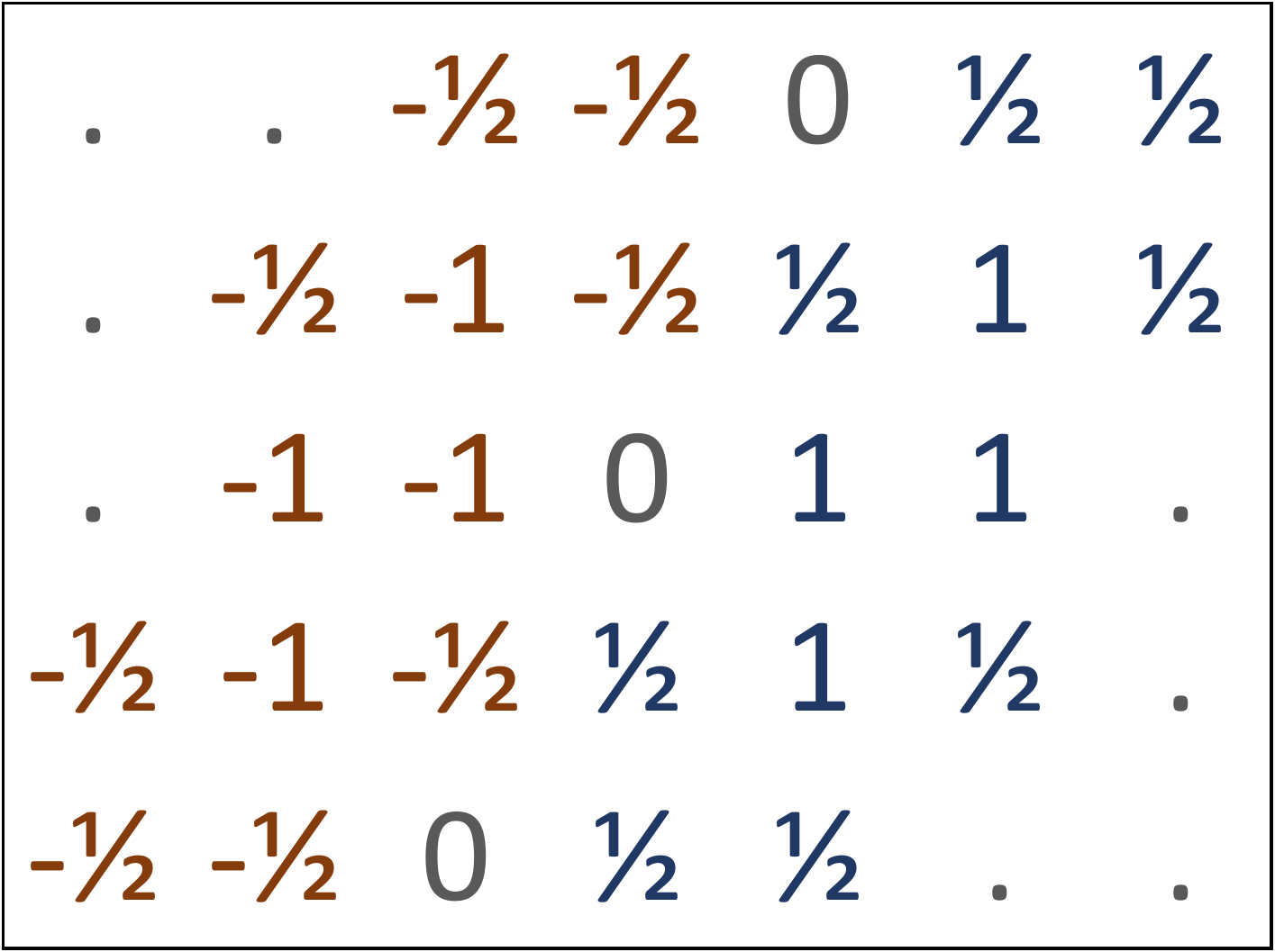}
\includegraphics[width=3.8cm,height=1.5cm]{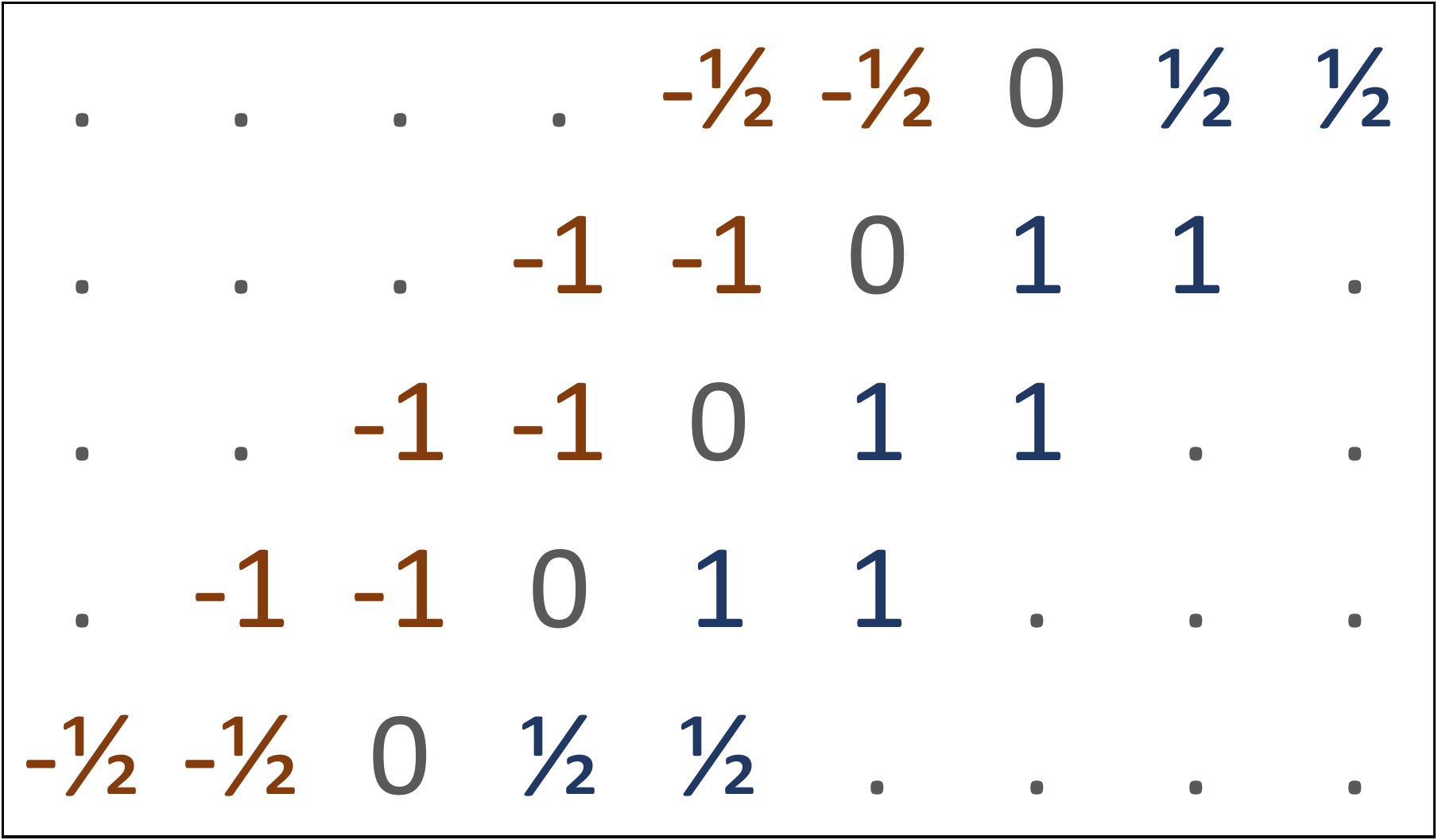}
\caption{Examples of filters of length $L=4$, width $w=4$, and spacing $s=1$ in clockwise offsets of (from left to right) $0^\circ$, $11.25^\circ$, $22.5^\circ$, and $45^\circ$ from vertical. The full set of vertical filters includes also a filter in orientation offset of $33.75$ and all of these filters reflected about the vertical axis. The horizontal filters are obtained by transposing the vertical ones. Our method computes the responses hierarchically so that the filters themselves are never explicitly constructed.}
\label{fig:Filters}
\end{figure}

Discrete versions of these filters are constructed by stacking shifted versions of a 1D, horizontal cross section filter in the vertical direction. Specifically, let $f_{cs} \in \mathbb{R}^{w+s}$ denote a cross section filter defined by concatenating the following three vectors, $(-1,\ldots,-1) \in \mathbb{R}^{w/2}$, $(0,\ldots,0) \in \mathbb{R}^s$, and $(1,\ldots,1) \in \mathbb{R}^{w/2}$. Each filter $f_{cs}$ is then duplicated at every row $0 \le y \le L$ and shifted horizontally by an offset of $y/\tan\theta$. Sub-pixel shifts are interpolated via second order interpolation. Lastly, the bottom and top rows are halved, to allow for a trapezoidal rule, and the filter is normalized by $wL$. Fig.~\ref{fig:Filters} shows several examples of these filters.

\begin{figure}
\begin{center}
\fbox{\includegraphics[width=2.7cm]{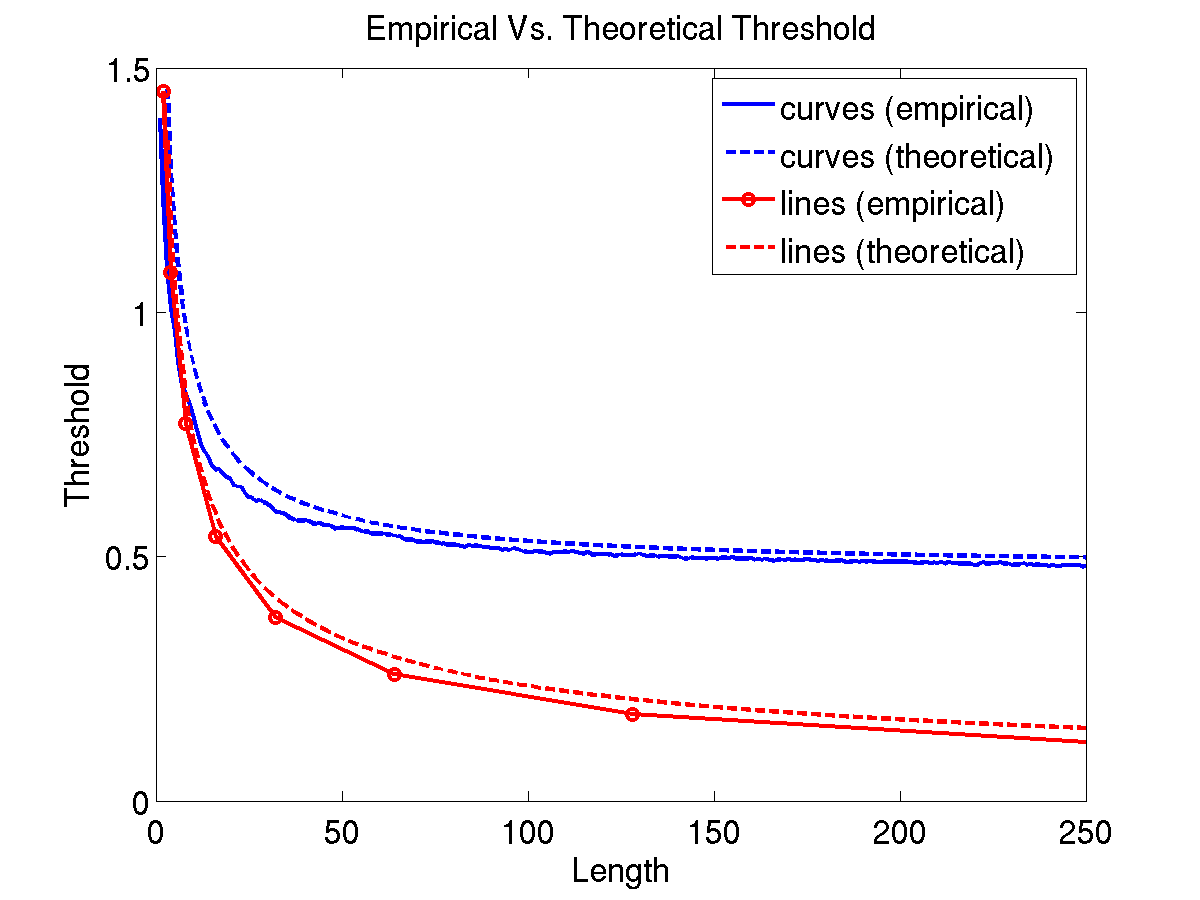}}
\end{center}
\caption{Detection threshold divided by the noise level $\sigma$, measured in SNR, as a function of length-$L$. The figure shows the theoretically derived threshold, i.e.,~\protect Eq. \eqref{eq:threshold} evaluated with $\delta=0.5,$  and an empirical evaluation of the median of the maximal responses obtained over a large set of pure noise images. The blue curves show the thresholds for the beam-curves binary tree. Both theoretical and empirical graphs approach a constant near 0.5. The red curves show the thresholds for straight lines, both approach zero.}

\label{fig:NoiseEstimation2}
\end{figure}

As we further explain in the next section, using our method, as we vary  $\theta$ and position, the total number of matched filters obtained for every length $L$ is roughly $K_L \approx 8 N$,  independent of $L$. The threshold function $T(L,K_L)$ of Eq.~\eqref{eq:threshold}, therefore, should decrease with $\sqrt{wL}$. This can indeed be seen in Fig.~\ref{fig:NoiseEstimation2}.
 where we show $T(L,K_L) / \sigma$ with $T(L,K_L)$ computed using~\eqref{eq:threshold} for an image of size $257 \times 257$ pixels and $w=4$. The figure also shows an empirical estimate of this ratio. This estimate is obtained by recording the maximal response over all length-$L$ responses, for $L=2,4,8,16,32,64,128,256$. It can be readily seen that the empirical estimates closely fit our expression.

%

\subsection{Hierarchical construction of filter responses}
\label{sec:framework}

We now describe an algorithm to compute the $O(N \log N)$ responses to the straight line filters defined in the previous section. Our method is based on~\cite{Dym99} and computes the responses hierarchically, so that \textit{the filters themselves are never explicitly constructed}. Below we describe the construction of the vertical responses (for which $\pi/4 \le |\theta| \le \pi/2$). For the horizontal responses, we simply transpose the image.


We first convolve the image with a single vertical cross section filter, $f_{cs}$, defined as before by concatenating $(-1,\ldots,-1) \in \mathbb{R}^{w/2}$, $(0,\ldots,0) \in \mathbb{R}^s$, and $(1,\ldots,1) \in \mathbb{R}^{w/2}$ and normalized by $w$. We denote the result of this convolution by $r(x_i,y_i)$ and refers to this as an array of {\em pixel responses}. We ignore pixels near the boundaries of the image for which $f_{cs}$ falls partly outside the image.

\begin{figure}[t]
\begin{center}
\begin{tabular}{|c|c|}
\hline\\[-4.1mm]
\includegraphics[width=0.08\linewidth]{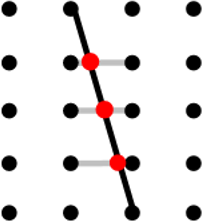}&
\includegraphics[width=0.30\linewidth]{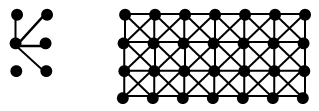}\\

\hline
\end{tabular}
\end{center}
\caption{Left panel:
Direct calculation of line integral ($L=4$): an interpolation from the given data points to the red
points is followed by the trapezoid rule. Right panel: Base level initialization: stencil of four length-1 responses for each pixel (left) cover the whole grid (right).}
\label{fig:baselevel}
\end{figure}

By definition, the family of filter responses are averages of $r(x_i,y_i)$ along line segments of different positions, lengths and orientations. For a line segment $\Gamma$ with endpoints  $(x_i,y_i)$ and $(x_j,y_j)$, its response $R(\Gamma)$ is the line integral of $r$ along $\Gamma$ and normalized by the  ($L_\infty$)
length of $\Gamma$. As illustrated in Fig \ref{fig:baselevel}(left), this response is  calculated numerically  by the trapezoidal rule and interpolation between horizontal neighbors.

As  mentioned earlier, the total number of line segments in an image with $N$ pixels is $O(N^2)$. However, following~\cite{Dym99} we calculate line integrals only over an $O(N \log N)$ subset of them. In particular, these responses are calculated hierarchically, and not directly. As we increase the  length \(L\), the spatial and angular resolutions change as follows:
\begin{itemize}
\item \textit{The minimal spatial resolution in the direction of
integration is inversely proportional to the integration length.}
In particular, when doubling the integration length, the number of
evaluation points is halved.
\item The spatial resolution \textit{perpendicular} to the direction of
integration is constant, independent of the integration length.
\item \textit{The minimal angular resolution is proportional to the
integration length.} In particular, when doubling the integration length, the number of angles computed is also
doubled.
\end{itemize}
A direct consequence of these principles is that the
\textit{total} number of integrals at any length $L$ is
independent of $L$.

\begin{figure}[t]
\begin{center}
\begin{tabular}{|c|c|}
\hline
\includegraphics[width=0.55\linewidth]{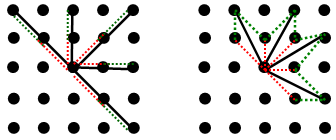}&
\includegraphics[width=0.22\linewidth]{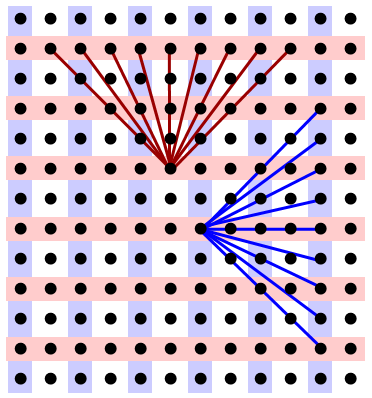}\\
\hline
\end{tabular}
\end{center}
\caption{Left panel: Integrals of length-2 are constructed from integrals
of length-1 as follows: i) for existing directions (left),  by averaging length-1 adjacent integrals (dashed lines) in that
direction and ii) for a new direction (right), by averaging four
nearest integrals (dashed lines). Right panel: The red lines
denote length-4 vertical responses, which are calculated at
every $2^{nd}$ row (pink rows), at each pixel in this row. The blue lines denote length-4
horizontal responses, which are calculated at every $2^{nd}$
column (blue columns), at each pixel in this column.} \label{fig:building}
\end{figure}

In the remainder of this section we follow~\cite{Dym99} and
describe a fast, \textit{hierarchical recursive calculation} of
``all significantly different'' filter responses in a discrete
image.
The hierarchical construction is as
follows. At the base level, for each pixel, four length-1 responses
are calculated (see Fig.~\ref{fig:baselevel}(right)). The
total number of responses at this level is therefore $4N$.
Recursively, given $4N$ filter responses of length $L$ we proceed to
computing new $4N$ responses of length $2L$. Following the
principles outlined above, the angular resolution should be
doubled. Consequently, the new responses can be divided into
two equally sized sets. Half of the new responses follow the
same directions as those of the previous level, while the other
half of the responses follow intermediate directions. The
first set of filter responses can be computed simply by taking the
average of two, length-$L$ responses with one coinciding
endpoint (see Fig.~\ref{fig:building}, left panel). The second
set of responses can be obtained by interpolation of four
length-$L$ responses of nearby directions. Each such set of four
responses forms a tight parallelogram around the desired
length-$2L$ integral. This is illustrated in
Fig.~\ref{fig:building} (left panel), where the average of the
four length-1 responses is used to construct a length-2
response. This can be viewed as first linearly interpolating
the two nearest directions to approximate the new direction at
length-1, then creating a length-2 response by averaging two
adjacent interpolated responses.

The total number of responses obtained with this procedure is $O(N \log N)$, and the procedure for computing them is linear in the number of output responses. While this is only a subset of the $O(N^2)$ possible straight responses, according to~\cite{Dym99} the remaining responses can be recovered by interpolation with a numerical error that is smaller than the error induced by the discretization of $I(x,y)$.



%

%

\subsection{Implementation}  \label{sec:straight_implementation}

Our implementation maintains a \textit{hierarchical data
structure} of the filter responses at multiple angular and spatial resolutions as follows. At
the base level, for each pixel the responses of a length-$1$ filters at four
orientations (vertical, horizontal and the two diagonals) are calculated. The angular resolution of length-$2L$ responses is twice the angular resolution of length-$L$
responses. The spatial resolution is halved as the length is
doubled as follows. Length-$L$ \textit{vertical} responses
($|{\theta}| \geq
\frac{\pi}{4}$) are calculated
at \textit{every $(L/2)^{th}$-row}, at each pixel in this row.
Length-$L$ \textit{horizontal} responses ($|{\theta}| \leq
\frac{\pi}{4}$) are calculated at \textit{every
$(L/2)^{th}$-column}, at each pixel in this column. In this
manner, each length-$L$ response has an \textit{overlap} of
length $L/2$ with another length-$L$ response of the same
orientation, in order to improve accuracy. As a result, at each
scale (length) the number of responses is $8N$, except for
the shortest level (length-$1$) for which $4N$ responses are
calculated. Note that at each length only $4N$ of the $8N$
responses are used to construct the subsequent level. The
grid setting for $L=4$ is illustrated in
Fig.~\ref{fig:building}(right).

During this recursive construction we compare the response $R(\Gamma)$ of every new candidate edge $\Gamma$ of length $L$ against the detection threshold of its respective length $T(L,K_L)$. In addition, we apply the consistent contrast test. For this test we need to first determine a consistency threshold using~\eqref{eq:consistent}, which requires an estimate of the mean $\mu_e$ and standard deviation $\sigma_e$ of the contrast along $\Gamma$. We estimate $\mu_e$ by the response $R(\Gamma)$ and $\sigma_e$ by the standard deviation of the intensities on each side of the edge. We note that we can efficiently compute these standard deviations during the recursive procedure by accumulating responses obtained by averaging the squared intensities $I^2(x_i,y_i)$ over the same set of positions, orientations, and lengths. Once we computed the consistency threshold for this test, we compare each of the sub-edges of $\Gamma$ against this threshold and accept $\Gamma$ only if the responses of all its sub-edges exceed the threshold. In practice, we thus need to compare only the sub-edge of minimal response against the threshold. This too can be done recursively by passing the minimal signed response from level to level, and, since we do not know in advance the sign of $R(\Gamma)$, also the maximal signed response.

\section{The Beam-Curve Binary Tree}  \label{sec:pyramid}

We next introduce a second edge detection algorithm. This efficient algorithm considers an exponentially sized subset of the set of general, non self-intersecting curved edges.

\subsection{Construction}

To detect statistically significant curved edges in an image \(I\), we first construct its \textit{beam-curve binary tree} and its corresponding data structure, denoted \(BC\).
To simplify the description we assume the image contains $N=n \times n$ pixels with $n=2^J+1$ for some integer $J$. We associate with our squared grid a system of square and rectangle tiles of different areas that are arranged in a binary-tree as follows. We use $j=0,1,2,...,j_m$ to denote scale. The tiles of different scales are aligned such that each tile $V$ at scale $j$ is subdivided into two sub-tiles, $V_1,V_2$ at scale $j+1$, where $V = V_1 \cup V_2$. The coarser level $j=0$ contains a single square tile of size $n\times n$. At the next level $j=1$ we split the square into two identical rectangles of size $n \times n/2$. At $j=2$ we split every rectangle into 2 identical squares of size $n/2 \times n/2$. In the next finer levels we continue to split in the same way recursively.

\begin{figure}[tb]
\centering
\includegraphics[width=3.4cm]{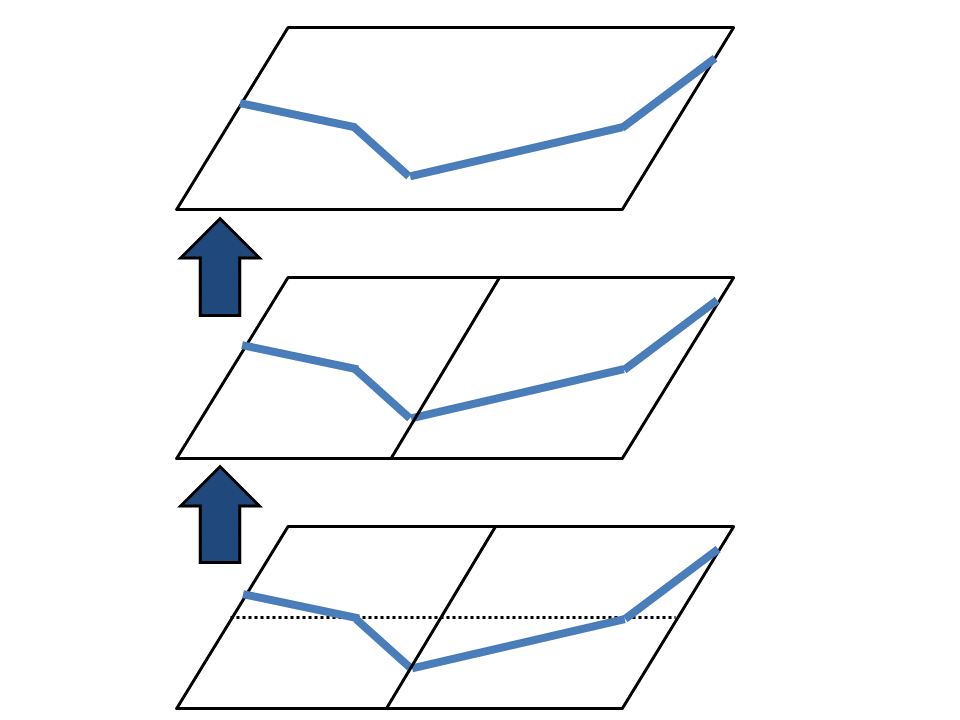}~~
\includegraphics[width=3.4cm]{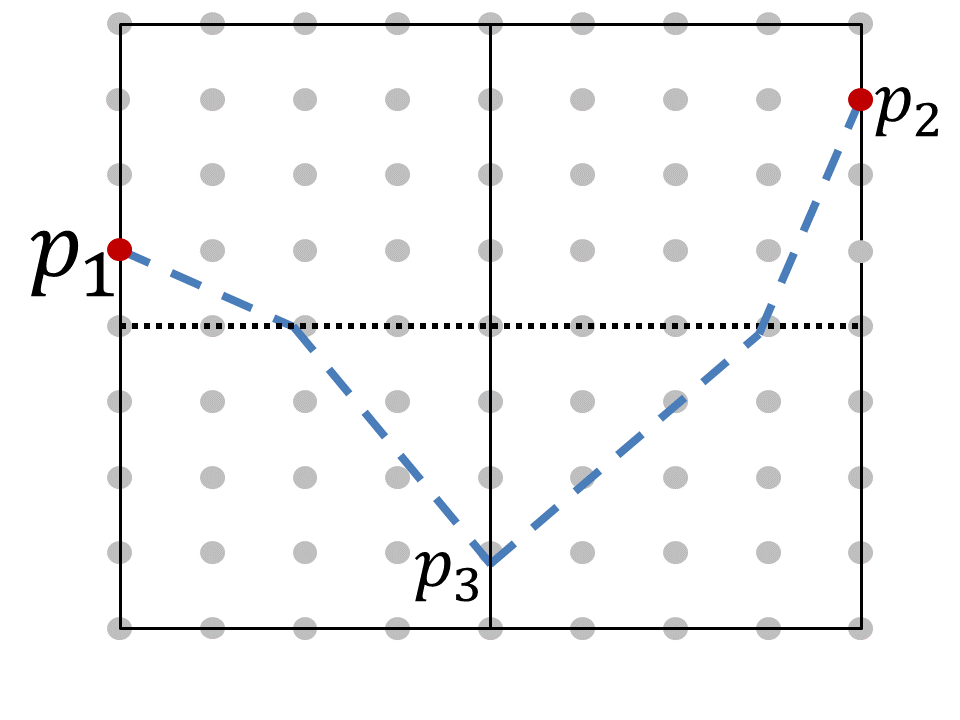}~~
\\[0.1cm]

\caption{Left: The three topmost levels of the beam-curve binary tree. Right: The $n\times n$ image at level $j=0$ is partitioned (solid line) into two rectangles of size $n \times n/2$ at level $j=1$. Each rectangle is then partitioned (dotted line) into two $n/2 \times n/2$ squares at level $j=2$.
A curve connecting two boundary pixels $p_1,p_2$ of level $j=0$ is a concatenation of up to 2 curves of level $j=1$, and up to 4 curves of level $j=2$.}
\label{fig:pyramid}
\end{figure}

Hence, at every even scale $j$ we cover the \(n\times n\) grid with a collection of square tiles of size $n/2^{0.5j} \times n/2^{0.5j}$ pixels such that each two sibling tiles share a common side (see Fig.~\ref{fig:pyramid}).
At every odd scale $j$ we cover it by a collection of rectangle tiles of size $n/2^{0.5j-0.5}\times n/2^{0.5j+0.5}$ with each two sibling tiles sharing one of their long sides. See Alg.~\ref{alg:main} for a pseudo-code of the tree construction

To each pair of points $p_1\in \partial V \cap \partial V_1$ and $p_2 \in \partial V \cap \partial V_2$ on the boundaries of different sides of a tile $V$ at level $j$ we associate a unique curve, referred to as {\em beam-curve}. At the finest scale $j=j_m$ the beam-curve is the straight line connecting $p_1$ and $p_2$. See Alg.~\ref{alg:bottom} for a pseudo-code of the bottom level processing.

At coarser scales $j<j_m$ the beam-curve connecting $p_1$ and $p_2$ is constructed recursively from beam-curves of scale $j+1$ as follows. Consider the collection of curves formed by concatenating curves from $p_1$ to any point $p_3 \in \partial V_1 \cup \partial V_2$ on the joint interface of $V_1$ and $V_2$, with curves connecting $p_3$ to $p_2$. We then select the curve with the highest filter response from this collection of curves. See Alg.~\ref{alg:merge} for pseudo-code of the coarser level construction.

\begin{algorithm}
\caption{$BeamCurveTree(V)$}
\label{alg:main}
\begin{algorithmic}
  \REQUIRE Tile $V$ whose maximal side length is $n$.
    \IF{$n\le n_{\min}$}
        \STATE $BC \gets BottomLevel(V)$
        \ELSE

    \STATE $V_1,V_2 \gets SubTiles(V)$
    \COMMENT{ The tile is split into two sub-tiles of equal area}
    \STATE $BC_1 \gets BeamCurveTree(V_1)$
    \STATE $BC_2 \gets BeamCurveTree(V_2)$
    \STATE $BC \gets CoarserLevel(V,V_1,V_2,BC_1,BC_2)$
    \ENDIF
    \RETURN $BC$
\end{algorithmic}
\end{algorithm}

\begin{algorithm}
\caption{$BottomLevel(V)$}
\label{alg:bottom}
\begin{algorithmic}
  \REQUIRE Small tile $V$.

        \STATE $BC \gets EmptySet$
        \FOR{ $\forall p_1,p_2 \in \partial V$}
                \STATE $\gamma \gets$ straight line from $p_1$ to $p_2$
                \STATE $BC.add(\phi(\gamma))$
        \ENDFOR
        \RETURN $BC$
\end{algorithmic}
\end{algorithm}

\begin{algorithm}
\caption{$CoarserLevel(V,V_1,V_2,BC_1,BC_2)$}
\label{alg:merge}
\begin{algorithmic}
  \REQUIRE $V$ is an image tile, $V_1$ and $V_2$ are its sub-tiles.
  \REQUIRE $BC_1$ is a set of the responses of sub-tile $V_1$.
  \REQUIRE $BC_2$ is a set of the responses of sub-tile $V_2$.

  \STATE $BC \gets$ $BC_1\cup BC_2$
  \IF{BasicMode}
  \STATE $InterfaceSet \gets \partial V_1 \cap \partial V_2$
  \ELSIF{OptimizedMode}
  \STATE $InterfaceSet \gets BestPixels(\partial V_1 \cap \partial V_2)$
  \ENDIF
        \FOR{$\forall p_1,p_2: p_1\in\partial V\cap \partial V_1,p_2\in\partial V\cap \partial V_2$}
             \STATE $AllResponses \gets EmptySet$
                    \FOR{$\forall p_3 \in InterfaceSet$}
                                \STATE $\gamma_1 \gets$ curve from $p_1$ to $p_3$ in set $BC_1$
                                \STATE $\gamma_2 \gets$ curve from $p_3$ to $p_2$ in set $BC_2$
                                \STATE $\phi(\gamma) \gets concatenate(\phi(\gamma_1),\phi(\gamma_2))$
                                \STATE $AllResponses.add(\phi(\gamma))$
                                \ENDFOR
                        \STATE $BC.add(AllResponses.bestResponse())$
        \ENDFOR
        \RETURN $BC$
\end{algorithmic}
\end{algorithm}

As we show in Section~\ref{sec:analysis}, the construction of the beam-curve pyramid allows us to search through an exponential set of curves, $K_L = O(N\cdot2^{0.65L})$. This set of curves is a much larger superset of the straight line segments used in both Section~\ref{sec:straight} and~\cite{Donoho02}. Still, beam-curves do not include various curves such as closed curves, spirals, and very windy curves\footnote{note that by a simple construction closed curves can be produced as well by this data structure. Specifically, given a tile of scale $j$, consider the cross shape produced by the boundaries between its descendant tiles. For every point $p$ on this cross we can produce closed curves by tracing a curve from that point through the four children sub-tiles and back to $p$.}. Our method represents such curves as a concatenation of (usually few) sub-curves.

While an exponential number of curves is scanned with this algorithm, the number of beam-curves stored in the pyramid and the cost of its construction are polynomial.
The number of beam-curves at every scale is roughly $6N,$ where $N$ denotes the number of pixels in the image (see Section~\ref{sec:analysis} below for details). The total number of beam curves therefore is $O(N \log N)$~\cite{Donoho02}. The cost of constructing the full pyramid of beam-curves is $O(N^{1.5})$ in a stringent version, and $O(N\log N)$ in a greedy version.
While this complexity may be high for certain practical applications, it can be reduced considerably by terminating the pyramid construction at a fixed scale or sparsifying the beam-curves through pruning. Speed up can also be gained by  a parallel implementation.

In Section~\ref{sec:analysis} we apply the analysis of Sec.~\ref{sec:threshold} to compute the detection thresholds in the beam-curve algorithm.
Since for beam-curves the algorithm searches through an exponential set of curves, the detection threshold tends to a strictly  positive constant for large images and edge lengths.



\textbf{A greedy version.} \label{sec:optimized}
For large images a computational complexity of $O(N^{1.5})$ may be prohibitive. Below we introduce a faster variant whose complexity is $O(N\log N)$, at the price of a slight decrease in  detection performance.

Let \(V\) be a tile at level \(j\) with children tiles \(V_{1},V_2\) and let $p_1, p_2 \in \partial V$. In this variant, the algorithm still looks for the curve with best response between \(p_{1},p_2\). However, instead of scanning all pixels in the joint interface $V_{12}=\partial V_1 \cap \partial V_2$, it only considers a subset of $k$ pixels for some fixed constant $k$. To select this subset, for each pixel $p_{3}\in V_{12}$ we look at the curve with highest response, that starts at either $\partial V_1$ or $\partial V_2$ and ends at $p_3$, as  previously computed at level $j+1$. We then keep only the $k$ pixels with highest responses. As we prove in Sec.~\ref{sec:analysis}, the overall number of operations of this variant is significantly smaller. Our experiments in Sec.~\ref{sec:experiments} indicate that empirically, this leads to a negligible decrease in edge detection quality.

\subsection{Implementation Details}

We start at level $j_m=\log_2(N)-4$ with tiles of size $5 \times 5$ pixels and associate a straight edge response with each pair of points on different sides of each tile. The mean intensity of a straight line $\gamma$ connecting two points $p_1$ and $p_2$ is
\begin{equation}
F(\gamma) = \frac{1}{L(\gamma)} \int_{p_1}^{p_2} I(p) dp,
\end{equation}
where we define the length as $L(\gamma) = \|p_2 - p_1\|_{\infty}$. As in the straight line case, we use the $\ell_\infty$ norm since it correctly accounts for the number of pixel measurements used to compute the mean. The mean is calculated by utilizing bi-cubic interpolation to achieve sub-pixel accuracy. We further calculate both $F$ and $L$ using the trapezoidal rule so that the end points are counted each with weight 1/2.

\begin{figure}[tb]
\centering
\includegraphics[height=1.8cm]{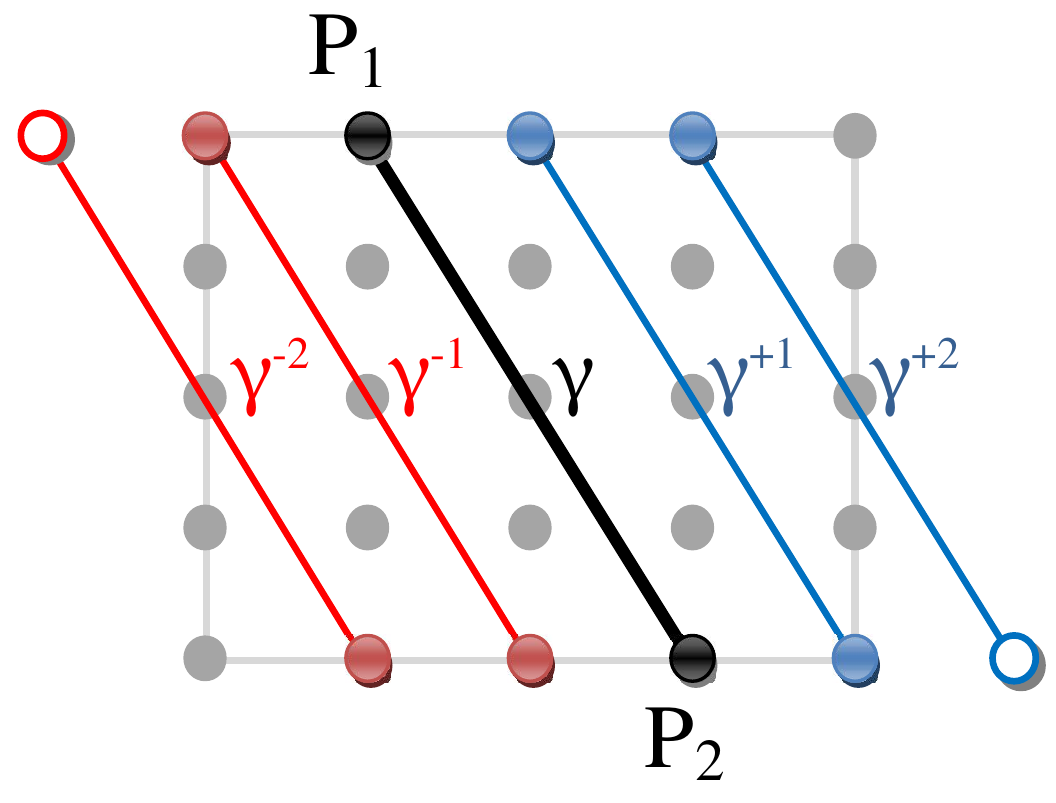}
\includegraphics[height=1.8cm]{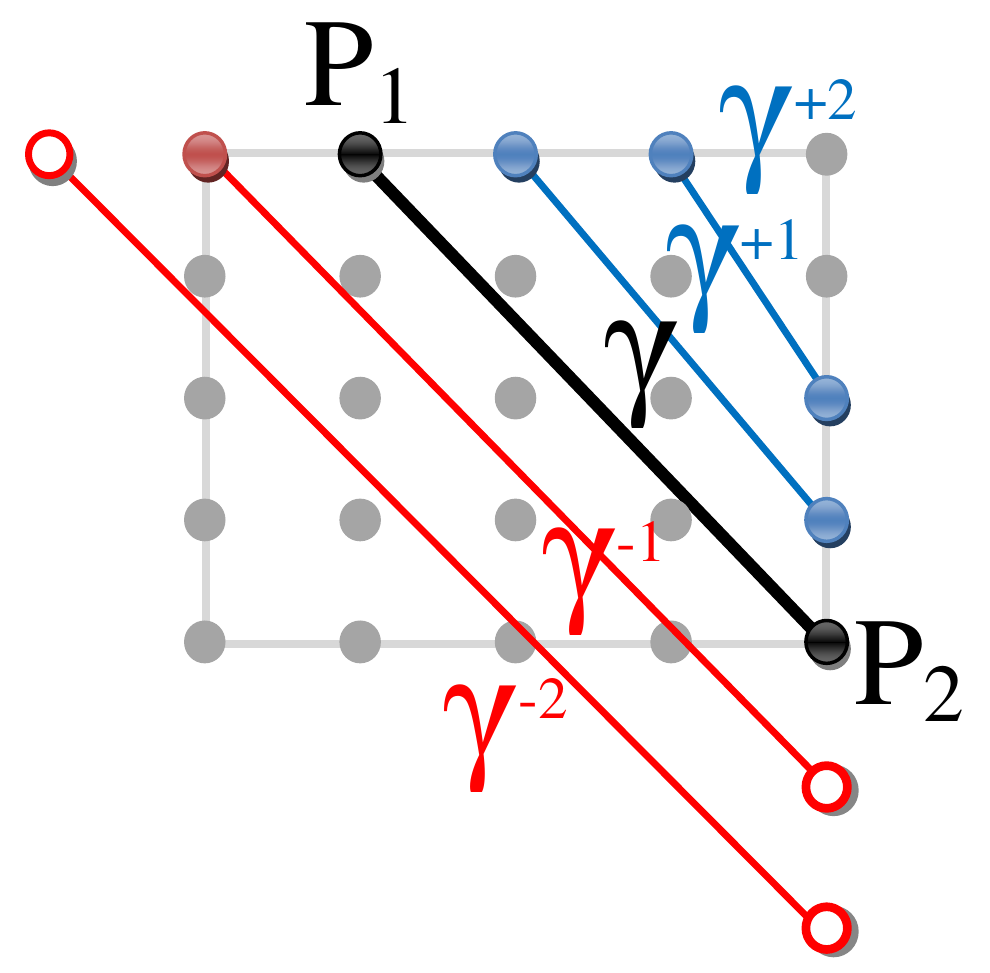}\\
\includegraphics[height=2.2cm]{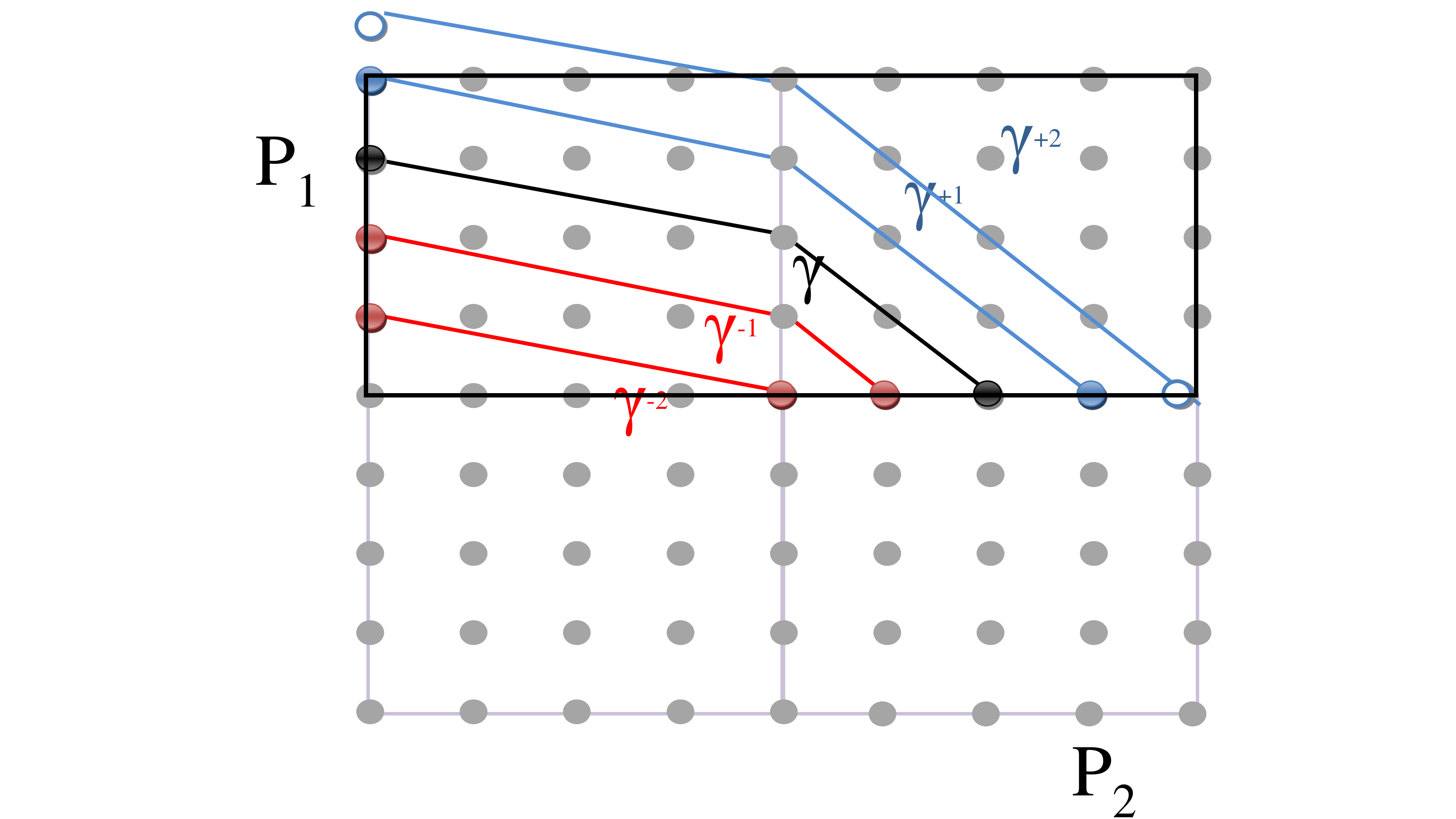}
\includegraphics[height=2.2cm]{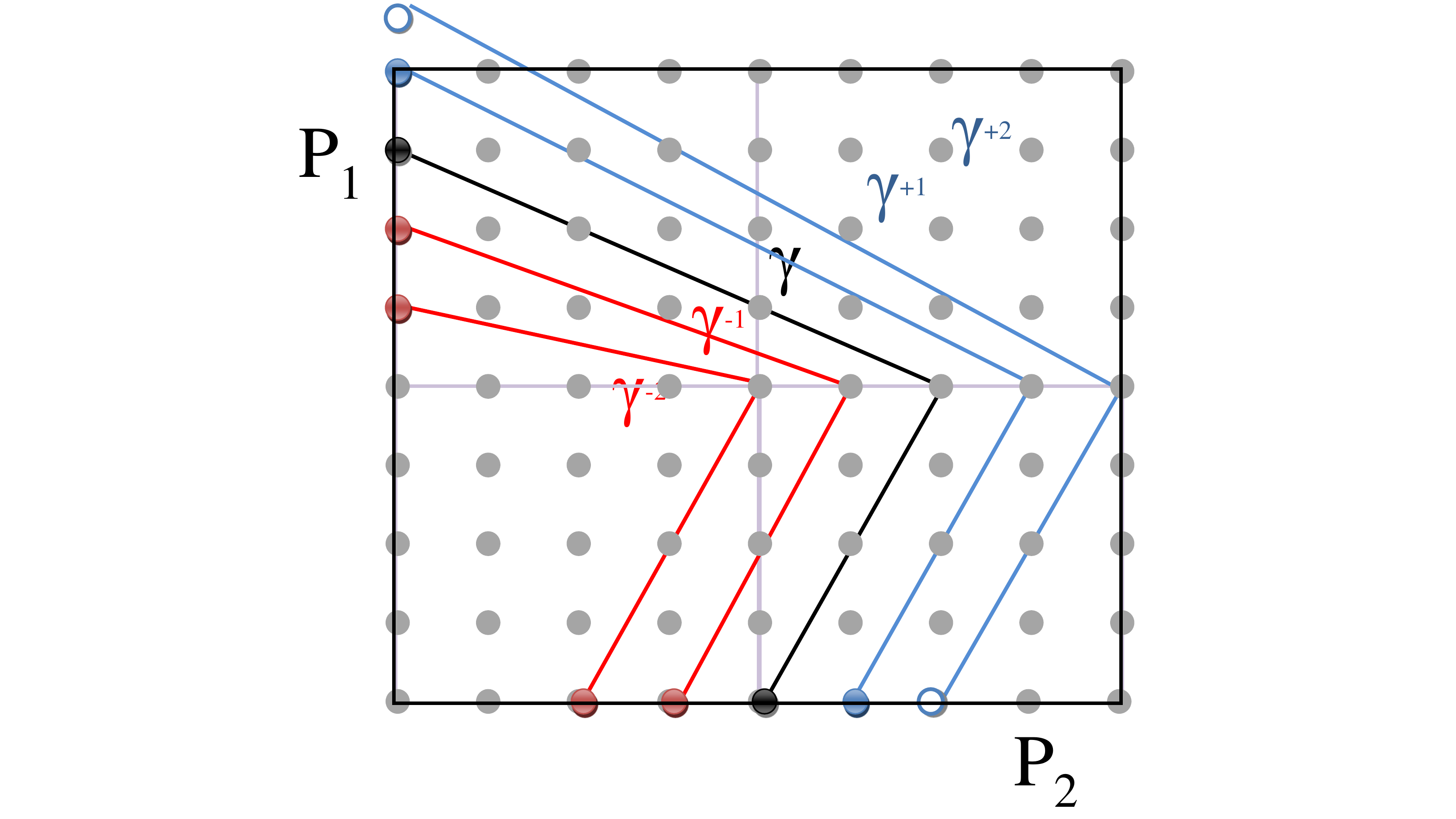}
\caption{Top row: Straight line filters of width $w=4$ in a $5 \times 5$ tile forming a parallelogram, and a general quadrangle. The offset curves may exceed beyond the boundaries of a tile. Bottom row:\ Stitching two straight filters at level $j_m$ to produce a curve at a rectangle of odd level $j_m-1$. (Right) Stitching two sub-curves of level $j_m-1$ to produce curves at a square of even level $j_m-2$. }
\label{fig:filters}
\end{figure}

Next, we define a response filter $R(\gamma)$, for a line $\gamma$ between $p_1$ and $p_2$ as follows. If $p_1$ and $p_2$ fall on opposite sides of a tile the filter has the shape of a parallelogram with
\begin{equation}  \label{eq:difference}
R(\gamma) = \left| \frac{\sum_{s=1}^{w/2} \left( L(\gamma^{+s}) F(\gamma^{+s}) - L(\gamma^{-s}) F(\gamma^{-s}) \right)}{\sum_{s=1}^{w/2} \left( L(\gamma^{+s}) + L(\gamma^{-s}) \right)} \right|,
\end{equation}
where $\gamma^s$ is the {\em offset line} connecting $p_1 + (s,0)$ with $p_2 + (s,0)$  if the points lie on a vertical side, or $p_1 + (0,s)$ and $p_2 + (0,s)$  in the horizontal case. Otherwise, if $p_1$ and $p_2$ fall respectively on horizontal and vertical sides the filter forms the shape of a general quadrangle (see Fig.~\ref{fig:filters}). The response is computed as in~\eqref{eq:difference}, where now the offset lines connect $p_1 + (s,0)$ with $p_2 \pm (0,s)$, depending on which of the four sides each of the points resides on. Note that the offset lines may fall partly outside a tile. In addition, every corner point is considered twice, once as lying on a horizontal side and once on a vertical side.

Once a level $j \le j_m$ in a pyramid is computed we proceed to construct the coarser level $j-1$. For each pair of points $p_1$ and $p_2$ on two different sides of a tile at level $j-1$ we consider all the curves that begin at $p_1$ and end at $p_2$ that can be obtained by stitching up to two curve segments of level $j$ while preserving continuity (see Fig.~\ref{fig:filters}). We then store for each such pair the curve that elicits the highest response. For the stitching we consider two curved segments $\gamma_1$ connecting $p_1$ with $p_3$ and $\gamma_2$ connecting $p_2$ with $p_3$ on two adjacent tiles at level $j$. We define the mean intensity of $\gamma = \gamma_1 \cup \gamma_2$ by
\begin{equation}  \label{eq:stitch}
F(\gamma) = \frac{1}{L(\gamma)} \left( L(\gamma_1) F(\gamma_1) + L(\gamma_2) F(\gamma_2) \right),
\end{equation}
where $L(\gamma) = L(\gamma_1) + L(\gamma_2)$. Note that due to the trapezoidal rule the point $p_2$ is counted exactly once. For the response we stitch the corresponding offset curves. Using~\eqref{eq:stitch}, we then compute their lengths and means and finally apply~\eqref{eq:difference} to obtain a response. Our recursion is similar but not indentical to dynamic programming,   since in some cases the optimal curve at a level $j-1$ may be composed of sub-optimal curves at level $j$.

During the construction of the beam curve pyramid we compare every new filter response of length $L$ against the detection threshold of its respective length $T(L,K_L)$ and apply the consistent contrast test, much like the case of straight edges (Section~\ref{sec:straight_implementation}). For this test we store with every filter response also its variance and  the minimal and maximal signed responses over its descendent curves.

\subsection{Analysis}  \label{sec:analysis}

\subsubsection{Computational complexity}

We now analyze the time and space complexity of our beam curve algorithm. Denote by $t(A)$ the number of operations performed by our algorithm on a tile $V$ of area $A$.
To compute all the responses of $V$ we scan over all triplets of pixels, $p_1\in\partial V_{1}$, $p_2\in\partial V_2$ and $p_3\in\partial V_1 \cap \partial V_2$ where  $V_1$ and $V_2$ the children tiles of $V$. Since the length of each of these three boundaries is $O(\sqrt A)$, the complexity of this step is proportional to $A^{1.5}$. This operation is repeated for the sub-tiles $V_1$ and $V_2$ whose areas are $\approx A/2$. Therefore, $t(A)$ satisfies the following recursion,
\begin{equation}  \label{eq:complexity}
t(A) = 2t(A/2)+O(A^{1.5}).
\end{equation}
The complexity $t(A)$ can be determined from the master theorem~\cite{master}, which considers recursions of the form
\begin{equation}  \label{eq:master}
t(n)=at(n/b)+f(n).
\end{equation}
The asymptotic behavior of $t(n)$ depends on the relation between  $f(n)$ and $n^{\log_b a \pm \epsilon}$. Specifically, if $f(n)=O(n^{\log_b a-\epsilon})$ for some constant $\epsilon > 0$ then $t(n)=\Theta(n^{\log_b a})$, while if $f(n)=\Omega(n^{\log_b a+\epsilon})$ for some constant $\epsilon > 0$ and in addition if $af(n/b)\le c f(n)$ for some constant $c < 1$ then $t(n)=\Theta(f(n))$. In \eqref{eq:complexity}, $a=b=2$ and $A^{1.5}=\Omega(A^{\log_2 2+0.5})$. In addition, $2f(A/2)=(1/\sqrt{2})f(A)$. Hence, $t(A)=\Theta(f(A))=\Theta(A^{1.5})$.
Finally, as the area of the root tile is the total number of image pixels $N$, the complexity of the beam-curve binary tree algorithm is $O(N^{1.5})$ operations.

Next we make a more refined analysis, including a derivation of the multiplicative constants. Recall that the number of pixels is $N = n \times n$ where $n = 2^J + 1$ for some integer $J \ge 0$. To simplify the analysis we assume that the finer level $j_m$ contains tiles of size $1\times1$, and therefore $j_m = \log_2 N$.  At every level $j$, $j=0,1,...,\log_2 N$, the image is partitioned into $2^j$ tiles. Every even level $j$ contains square tiles of size $n/2^{0.5j} \times n/2^{0.5j}$, while every odd level $j$ contains rectangle tiles of size $n/2^{0.5j-0.5} \times n/2^{0.5j+0.5}$.
Therefore, the total number of beam curves considered in any tile of an even level $j$ is,
\begin{equation}
{4 \choose 2} \times n/2^{0.5j} \times n/2^{0.5j} = 6N/2^j.
\end{equation}
Since there are $2^j$ tiles in level $j$, the total number of curves at this level is $6N$.
It can be shown that the number of curves in every odd level $j$ is the same as in every even level. Thus, the total number of beam curves summing over all levels is roughly $6N \log N$.

Next we analyze the number of operations to construct the beam curves. At every scale $j$ we construct a new curve by connecting two sub-curves of level $j+1$. The number of considered stitchings equals the length of the joint interface. This length equals $n/2^{0.5j}$ at any even level $j$, and $n/2^{0.5j+0.5}$ at any odd level. Thus, the total number of beam curves at every level $j$ is $6N$, times the interface length at even levels,
\begin{equation}
6N \times n/2^{0.5j} = 6N^{1.5} /2^{0.5j},
\end{equation}
and at odd levels,
\begin{equation}
6N \times n/2^{0.5j+0.5} = 6N^{1.5} /2^{0.5j+0.5}.
\end{equation}
Summing these over all scales we get the following complexity, denoted $C(N)$, \begin{equation}
C(N)=6N^{1.5}\Bigg[\sum_{j_{even}}^{\log_2 N}2^{-0.5j}+\sum_{j_{odd}}^{\log_2 N}2^{-(0.5j+0.5)}\Bigg].
\end{equation}
Consequently,
\begin{equation}
C(N) \lesssim 6N^{1.5}\Big[\sum_{k=0}^{\infty}2^{-k}+\sum_{k=1}^{\infty}2^{-k}\Big]=18N^{1.5}.
\end{equation}

\textbf{Greedy Algorithm Complexity.} As described in Sec.~\ref{sec:optimized}, the faster algorithm scans, for each of the $6N$ pairs of start and end points, only the top $k$ pixels in the joint interface. Overall, it considers $6Nk$ curves at each level $j$. Additional work is required to select the best $k$ pixels. As the number of tiles is $2^j$ and the interface length is $\approx n/2^{0.5 j}$, this step makes at most $(2^j n/2^{0.5 j})\log k<N$ operations for every level $j$. To conclude, the number of operations at each level is at most $(6k+1)N.$ Thus,
the overall complexity is bounded by $(6k+1)N\log N$, which leads to a significantly faster runtime.

\subsubsection{Detection thresholds}

Next we apply our analysis of Sec.~\ref{sec:threshold} to compute the detection thresholds for the beam-curve algorithm. We begin by computing the size of the search space, $K_{L}$, of candidate curves of length $L$ in the beam-curve binary tree (BCBT). This quantity directly affects the contrast threshold~\eqref{eq:threshold}.

We first calculate the search space size at level $j$ of the BCBT, and then show its connection to $K_L$. Denote by $S(j)$ the upper bound of the total number of candidate curves at level $j$, and denote by $s(j)$ the same number, but for given fixed start and end points. Since  the total number of stored curves at any level is approximately $6N$, then
\begin{equation} \label{eq:S_S_tag}
S(j) = 6N s(j).
\end{equation}
Next, to compute $s(j)$, recall that in the BCBT, we split a tile $V$ at level $j$ into two sub-tiles $V_1,V_2$ at level $j+1$, with a joint interface $\partial V_1 \cap \partial V_2$ of length $\approx n/2^{0.5 j}$. For fixed endpoints $p_1,p_2\in\partial V$, the quantity $s(j)$ satisfies the following recursive formula
\begin{equation}
s(j) = s^2(j+1) n/2^{0.5 j}.
\end{equation}
To apply the master theorem~\cite{master}, take logarithm on both sides
\begin{equation}
\log s(j) = 2\log s(j+1)+ \frac{1}{2}\log (N/2^{j}).
\end{equation}
Substitute $A=N/2^{j}$, the number of pixels in a tile at level $j$, and define $\tilde s(A) = s(j)$. Then,
\begin{equation}
\log \tilde s(A) = 2\log \tilde s(A/2)+ \frac{1}{2}\log A.
\end{equation}
According to~\eqref{eq:master}, denote $t(A)=\log \tilde s(A)$ and $f(A)=0.5\log A$. Therefore, in this case $a=b=2$ and $f(A)=O(A^{\log_2 2-0.5})$. By the master theorem $t(A)=\log \tilde s(A)=\Theta(A)$. Subsequently, $s(j) =  2^{O(N/2^{j})}$ and combining this with Eq.~\eqref{eq:S_S_tag},
\begin{equation} \label{eq:sp}
S(j) =  (6N) 2^{O(N/2^{j})}.
\end{equation}
To derive an expression for $K_L$, it can be shown, that the average length over all candidate curves in a given tile at level $j$ is proportional to the area of the tile, $L =  O(N/2^{j}).$  Therefore we can approximate $K_L$, by
\begin{equation} \label{eq:kl}
K_L \approx (6N) 2^{\beta L}
\end{equation}
for some constant $\beta$. We found by empirical fitting that $\beta \approx 0.65$.

Inserting $K_L$ in the expression for the threshold~\eqref{eq:threshold} gives
\begin{equation} \label{eq:threshold2}
T(L) = \sigma \sqrt{\frac{2 \ln({6N \cdot 2^{\beta L})}}{wL}}.
\end{equation}

\textbf{Minimal Detectable Contrast.} An interesting question, already raised above, is how faint can an edge be and still be detected. In our case, note that as $N$ and $L$ tend to infinity, the threshold in Eq.~\eqref{eq:threshold2} converges to a strictly positive limit,
\begin{equation}
T_\infty = \Omega\Big(\frac{\sigma}{\sqrt{w}}\Big).
\end{equation}
Namely, due to the exponential size of the search space, our threshold is bounded from below by a positive constant. Hence, our ability to detect faint edges of unknown shape and location in low SNR is limited. Fig.~\ref{fig:NoiseEstimation2} compares the theoretical threshold of Eq.~\eqref{eq:threshold2} to empirical results, for both the beam curve and straight line algorithms. It can be seen that both curves are close to each other, and that the graphs converge to $\approx 1/2$. This value is the asymptotic bound $T_\infty$ for the selected parameters in this simulation: width $w = 4$, image size $N = 129 \times 129$, noise level $\sigma = 1$ and $\beta = 0.65$.

\section{Non-maximum suppression}

Typically, real edges also give rise to responses that exceed the threshold at adjacent locations and orientations.
To output well-localized edges expressed by a single response, we conclude each algorithm by non-maximal suppression.

For the line detection algorithm, during the bottom-up line construction we perform at each length $L$ \textit{angular} followed by \textit{spatial} non-maximal suppression. For angular suppression, for each set of lines of a certain length $L$ whose centers coincide we discard all lines except for the line of highest response. Then, for the remaining lines at length $L$  the spatial suppression scan keeps the lines of the highest response with respect to their spatial neighbors.
We conclude the non-maximum suppression part by applying inter-level non-maximum suppression. We perform this suppression to ensure that each edge is detected at its maximal length. We begin by accepting all the lines of maximal possible length $L_{\mathrm{max}}$ that survived the previous steps. Then, recursively, assuming we performed the suppression of lengths $2L$ and up, we apply suppression to lines of length $L$ as follows. For each line of length $L$ we measure its amount of overlap with accepted lines of length $2L$. Any line whose measured overlap exceeds a prescribed fraction (we used 0.52) is removed. For lines with smaller overlap than the prescribed fraction, the non-overlapping portion of the line, say of length $\tilde L$, is tested against the appropriate detection threshold $T(\tilde L,K_{\tilde L})$ \eqref{eq:threshold} and is accepted if it exceeds this threshold.


The non-maximum suppression for the curved edge detection algorithm is similar and is explained next. Starting from the top-most tile in the pyramid, in each tile we process the curves in a descending order of their responses. For each curve, we accept it if its offset curves do not overlap with previously selected curves. If however they partially overlap with the already accepted curves we discard the overlapping portion and accept the remaining portion if it exceeds the appropriate detection threshold. To test if a curve overlaps with the already accepted curves we consider only its sub-curves at all levels below and declare an overlap when its symmetric median Hausdorff distance to an already accepted curve falls below a given threshold (we used 1-3 pixels, depending on the width of the filter).




\begin{figure}\label{fig:cc}
\centering

\fbox{\includegraphics[width=1.4cm]{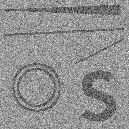}}
\fbox{\includegraphics[width=1.4cm]{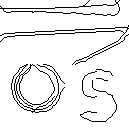}}
\fbox{\includegraphics[width=1.4cm]{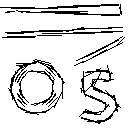}}
\fbox{\includegraphics[width=1.4cm]{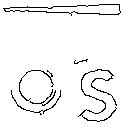}}
\fbox{\includegraphics[width=1.4cm]{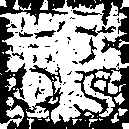}}
\\[1mm]
\fbox{\includegraphics[width=1.4cm]{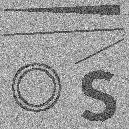}}
\fbox{\includegraphics[width=1.4cm]{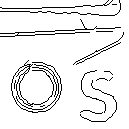}}
\fbox{\includegraphics[width=1.4cm]{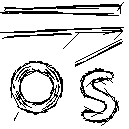}}
\fbox{\includegraphics[width=1.4cm]{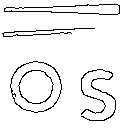}}
\fbox{\includegraphics[width=1.4cm]{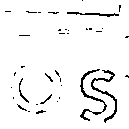}}
\caption{Result of  various edge detection algorithms to the noisy simulation image, at SNR 2 (top), and SNR 3 (bottom). From left to right: Input image, our curves $O(N^{1.5})$, our lines $O(N\log N)$, Canny and PMI.}
\label{fig:simulations}
\end{figure}

\section{Experiments} \label{sec:experiments}

\begin{figure}
\begin{center}
\includegraphics[width = 0.24\textwidth]{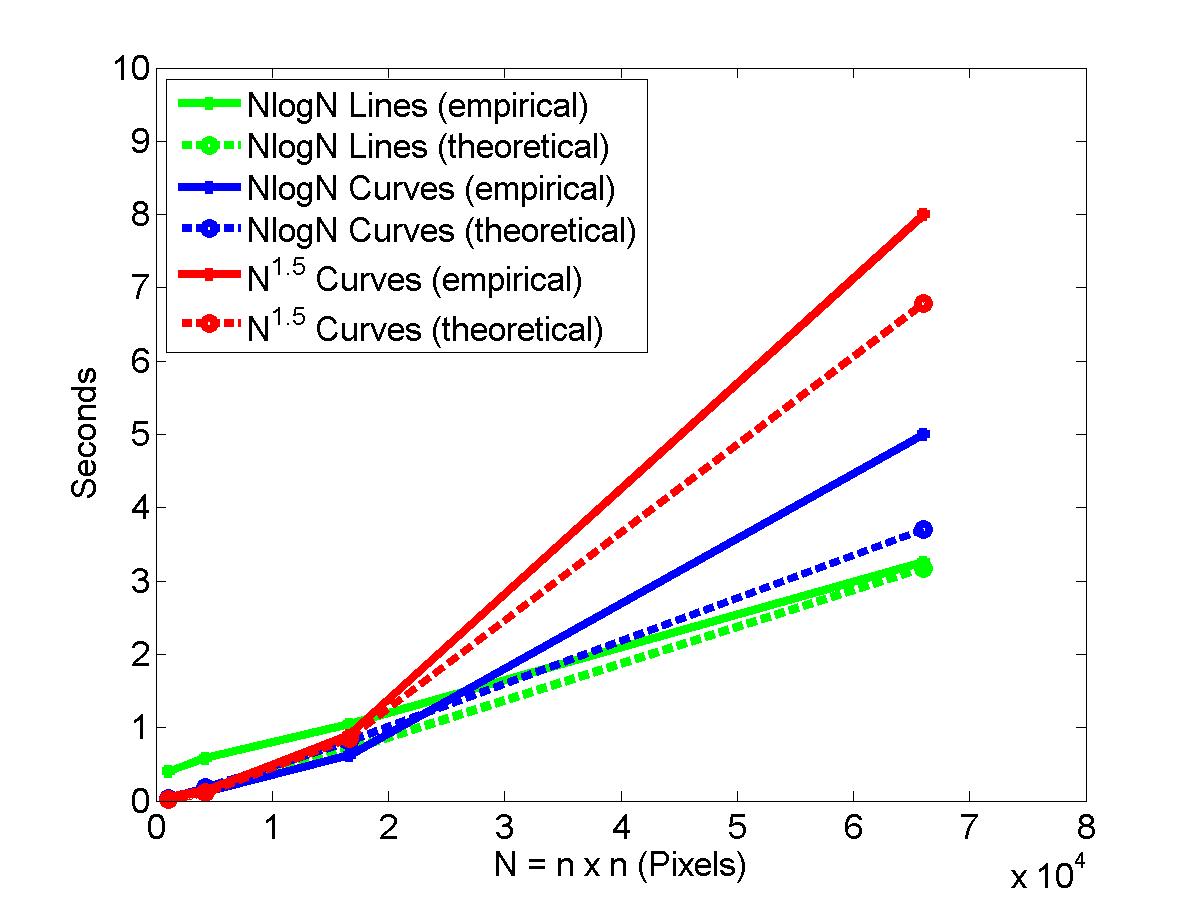}
\includegraphics[width=0.24\textwidth]{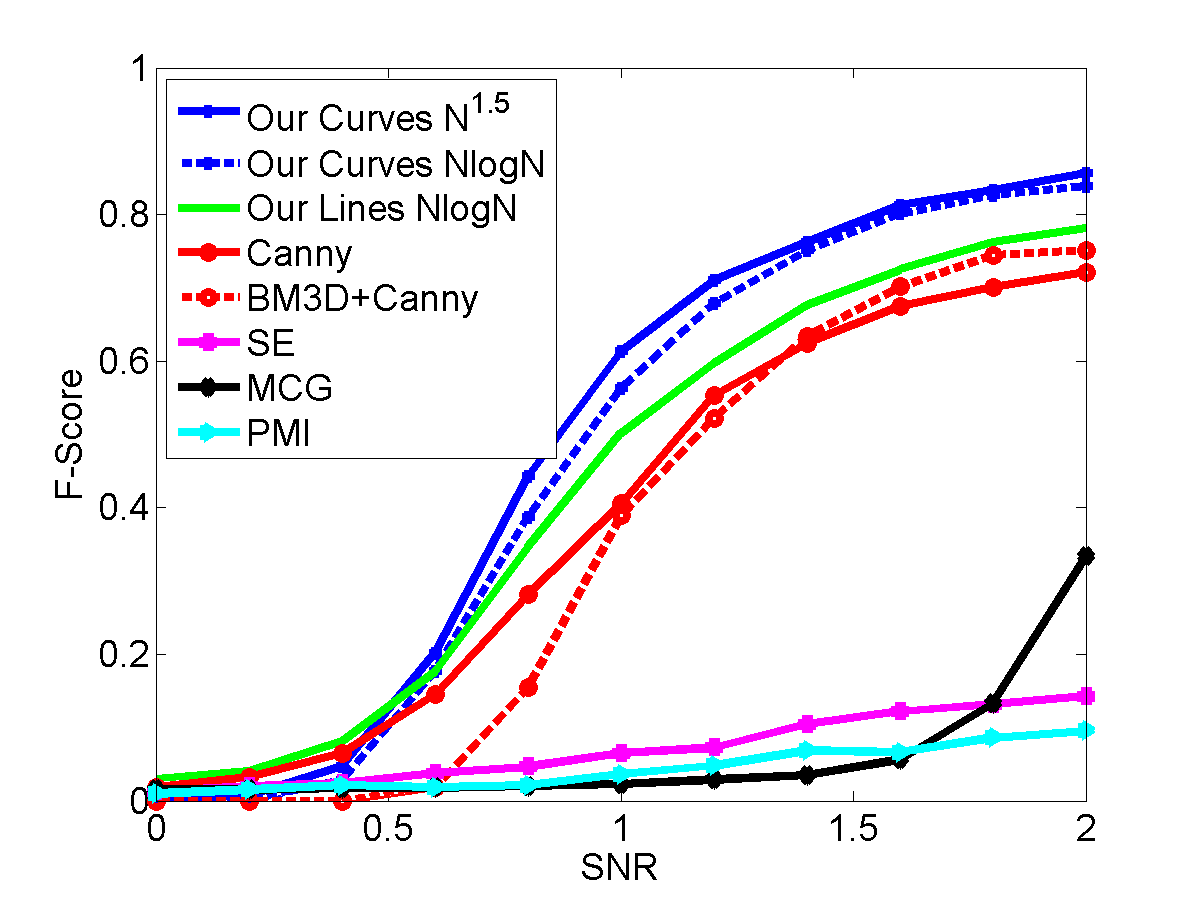}
\end{center}
\caption{(Left): Empirical run-time of our straight and curved edge detectors compared with their theoretical run-times.
(Right) Simulation results: F-measures of various edge detection algorithms as a function of SNR.}
\label{fig:runtime}
\end{figure}

We implemented our straight edge detector (denoted Lines) in Matlab and our curved edge detector (denoted Curves) in C++. Furthermore, as our tree construction can be easily parallelized, our implementation of Curves utilizes multi-threading. We ran our experiments on a single 8-core Intel i7, 16 GB RAM machine. For an input image with $129\times 129$ pixels Lines runs in $\approx 1$ second, while Curves C++ is $\approx 0.6$ second for the $O(N\log N)$ version with $k=40$ and $\approx 0.9$ second for the $O(N^{1.5})$. For an image with $257\times 257$ pixels the run-times respectively are $\approx 3$, $5$ and $8$ seconds. These run times include the post processing step of computing the final edge map image. More CPU cores can reduce these run times significantly. The runtime graphs are shown in Fig.~\ref{fig:runtime}.

In our experiments, we considered both challenging simulated images as well as to real images acquired under unfavorable photography conditions.
 We compared our Lines and Curves detectors to the classical Canny~\cite{Canny86} algorithm, and to several state of the art algorithms for boundary detection in natural images, including Multiscale-Combinatorial-Grouping (MCG)~\cite{MCG}, Crisp Pointwise-Mutual-Information (PMI)~\cite{Crisp} and Structured-Edges (SE)~\cite{Dollar}.

\subsection{Simulations}
We prepared a binary pattern of size $129 \times 129$ pixels 
containing several shapes, including two concentric circular rings of widths 2, two long straight bars of width 2, a single long triangle and the letter 'S'. We next scaled the intensities in each pattern by a factor $\tau$ and added i.i.d.~zero mean Gaussian noise with standard deviation $\sigma$, thus producing images with SNR $\tau / \sigma$. For each binary pattern we produced 11 images with SNRs ranging between 0 and 2.


We evaluate the results by the F-measure $F=2PR/(P+R)$, which is the harmonic mean of precision $P$ and recall $R$ ~\cite{vanrijsbergen79information}. Fig.~\ref{fig:simulations} shows the noisy simulation pattern with SNRs 2 and 3 along with detection results for the various algorithms. With this SNR our algorithms managed to detect nearly all the edges with very few false positives. The results for SNRs 0-2 are summarized in Fig.~\ref{fig:runtime} (right)
and Table~\ref{tab:F-measure}.
All methods are tuned to detect no edges when the image includes pure noise (SNR=0). For that sake, we modified  Canny's thresholds to  $[low,high] = [0.28,0.7]$. For the other methods, we used a fixed threshold on the soft edge map.
It can be seen that our methods significantly outperform the compared algorithms. It should be noted however that some of the compared algorithms are not designed specifically to handle significant amounts of noise. Evidently, algorithms that achieve state-of-the-art results on natural images do not cope well with noisy images.

%

\begin{table}[h]
\caption{Average F-measures in simulations over images at three SNR ranges.}
\begin{center}
\begin{scriptsize}
\begin{tabular}{|l|c|c|c|}
\hline
\textbf{Algorithm $\backslash$ SNR Range}  &  \textbf{0.0-0.8} & \textbf{1.0-1.6} & \textbf{1.8-2.0}  \\
\hline
\textbf{Our Curves $O(N^{1.5})$} & \textbf{0.14} & \textbf{0.73} & \textbf{0.85} \\
\hline
\textbf{Our Curves $O(N\log N)$} & \textbf{0.12} & \textbf{0.7} & \textbf{0.83} \\
\hline
\textbf{Our Lines} & \textbf{0.14} & \textbf{0.63} & \textbf{0.77} \\
\hline
Canny   & 0.11 & 0.57 & 0.71 \\
\hline
BM3D+Canny  & 0.04 & 0.57 & 0.75 \\
\hline
SE  & 0.03 & 0.09 & 0.14 \\
\hline
MCG & 0.02 & 0.04 & 0.23 \\
\hline
PMI & 0.02 & 0.06 & 0.09 \\
\hline
\end{tabular}
\end{scriptsize}
\end{center}
\label{tab:F-measure}
\end{table}

\subsection{Real Images}

\begin{figure*}
\begin{center}
\fbox{\includegraphics[width=1.6cm]{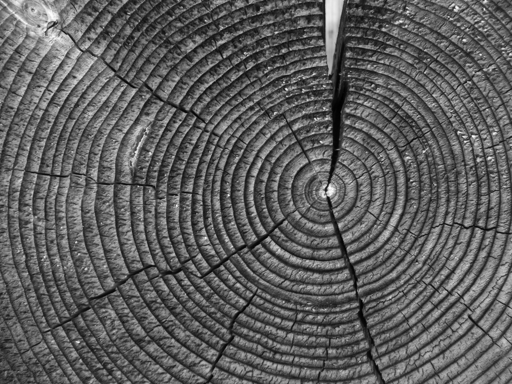}}
\fbox{\includegraphics[width=1.7cm]{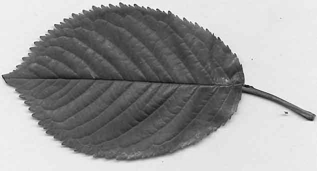}}
\fbox{\includegraphics[width=1.6cm]{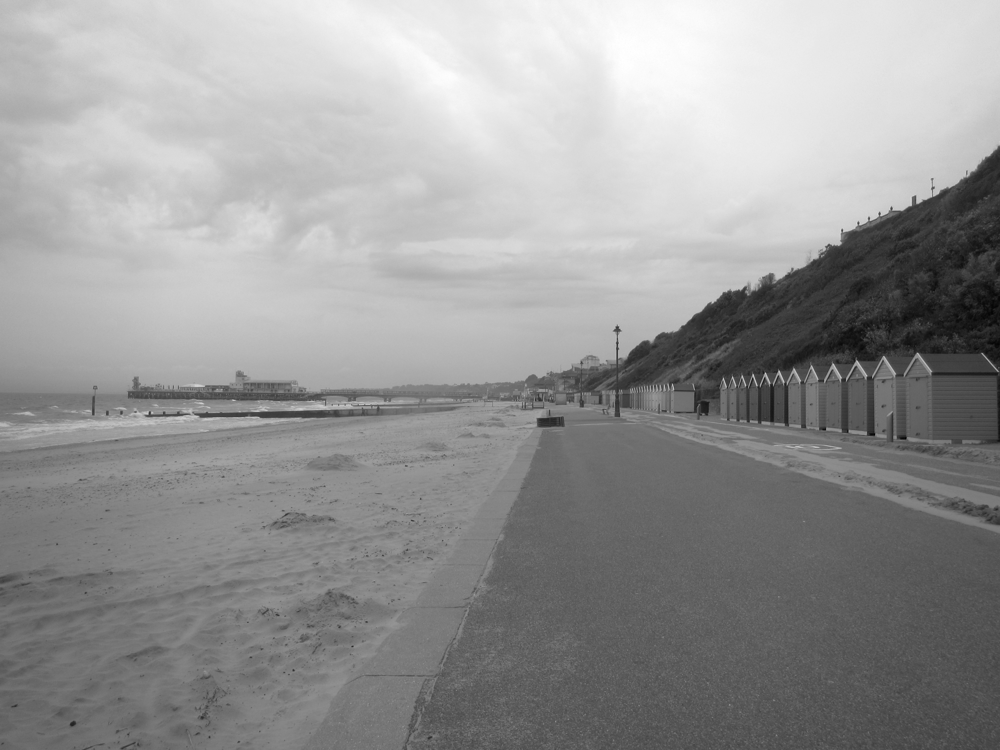}}
\fbox{\includegraphics[width=2cm]{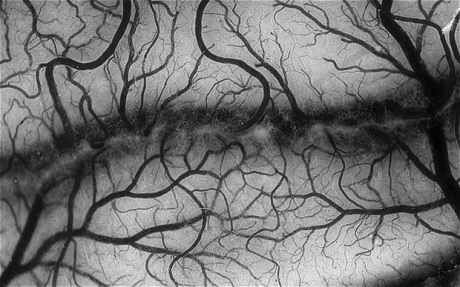}}
\fbox{\includegraphics[width=1.6cm]{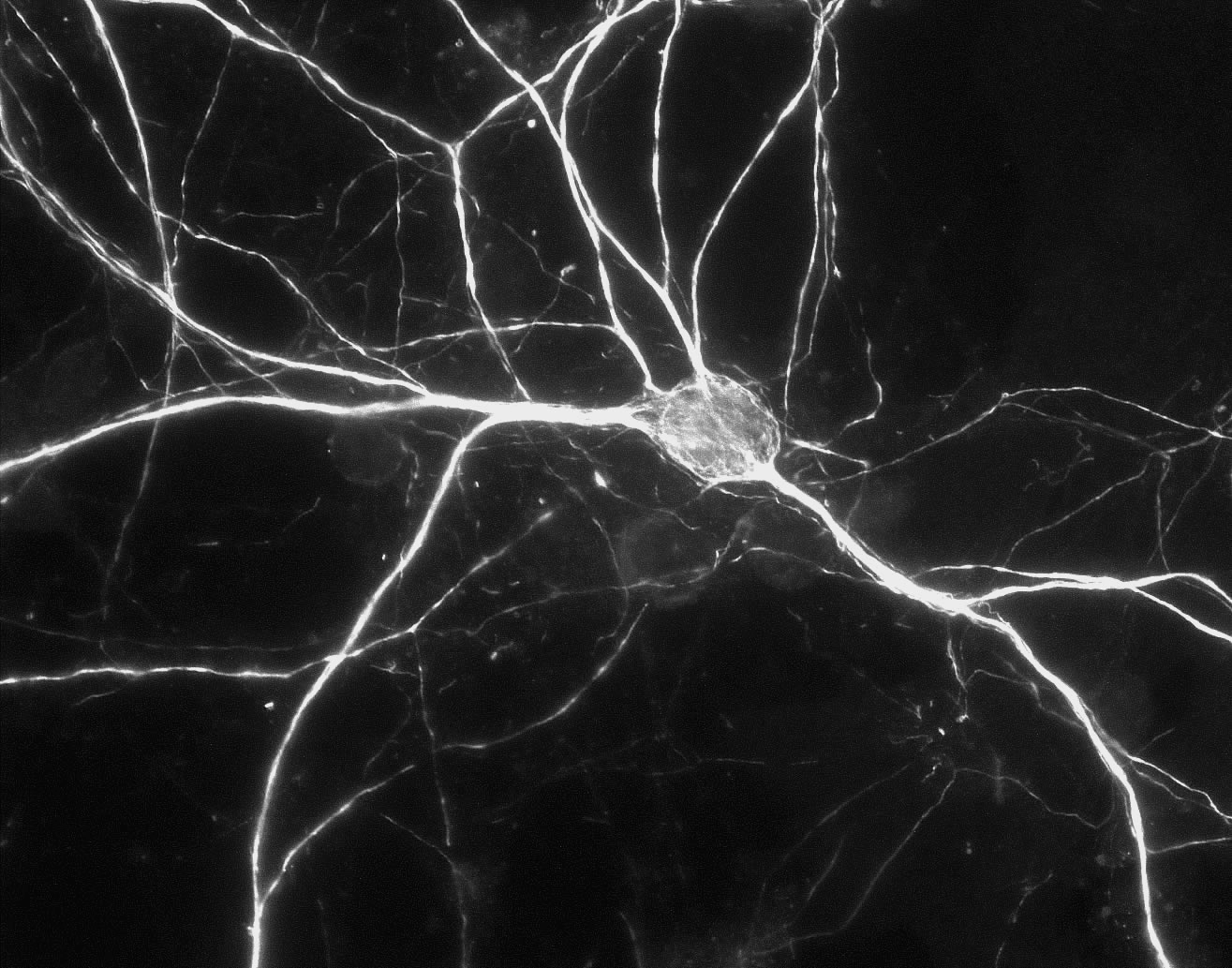}}
\fbox{\includegraphics[width=1.8cm]{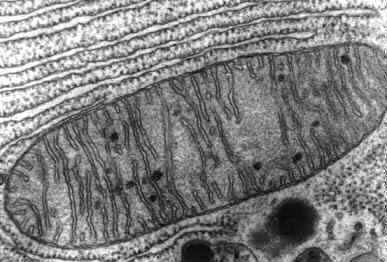}}
\fbox{\includegraphics[width=1.7cm]{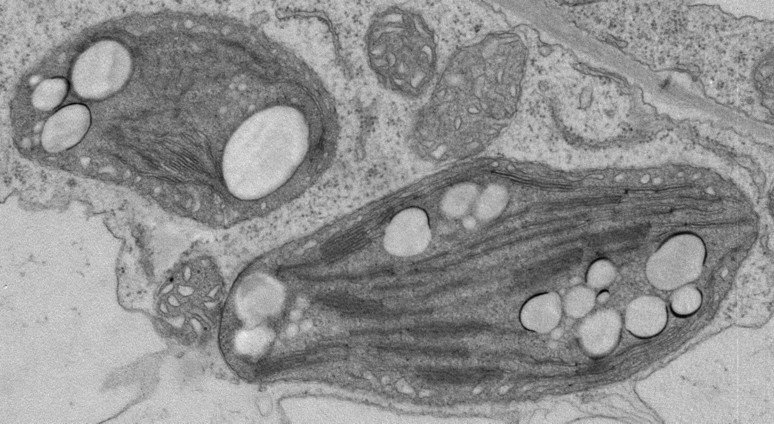}}
\fbox{\includegraphics[width=1.8cm]{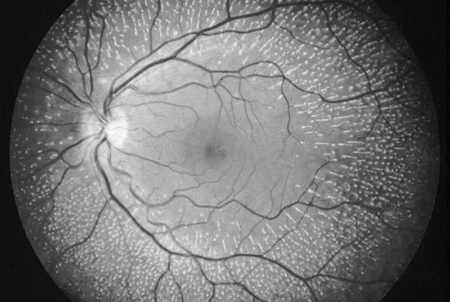}}
\fbox{\includegraphics[width=1.7cm]{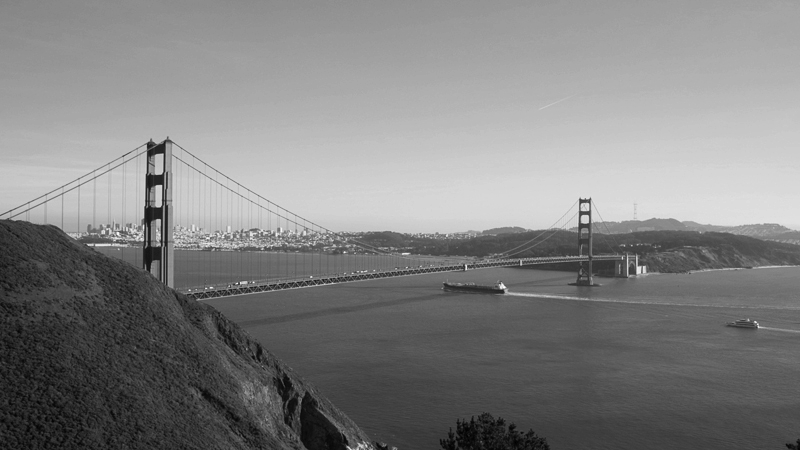}}
\\[0.1cm]
\fbox{\includegraphics[width=1.6cm]{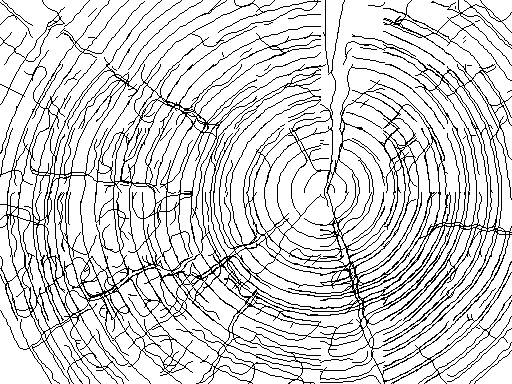}}
\fbox{\includegraphics[width=1.7cm]{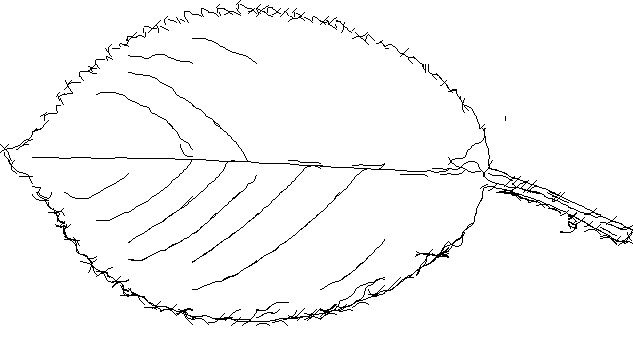}}
\fbox{\includegraphics[width=1.6cm]{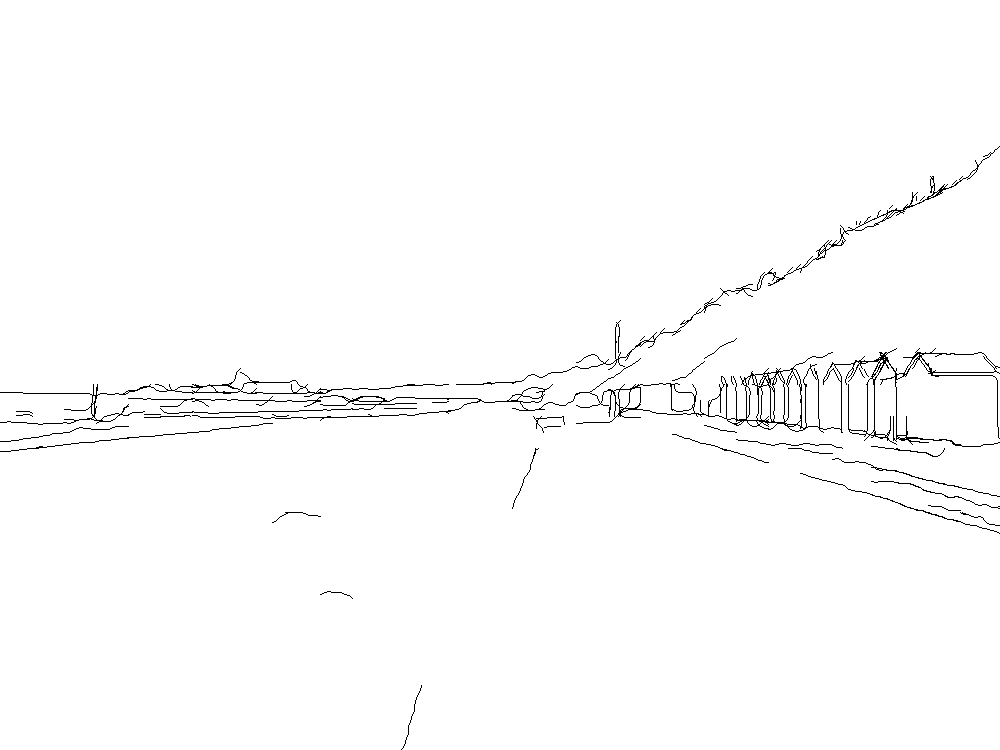}}
\fbox{\includegraphics[width=2cm]{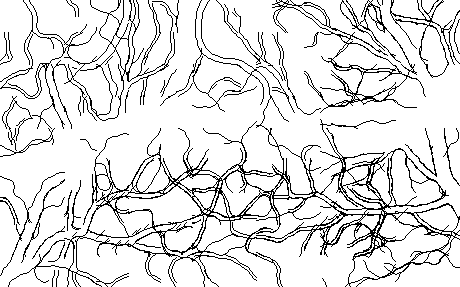}}
\fbox{\includegraphics[width=1.6cm]{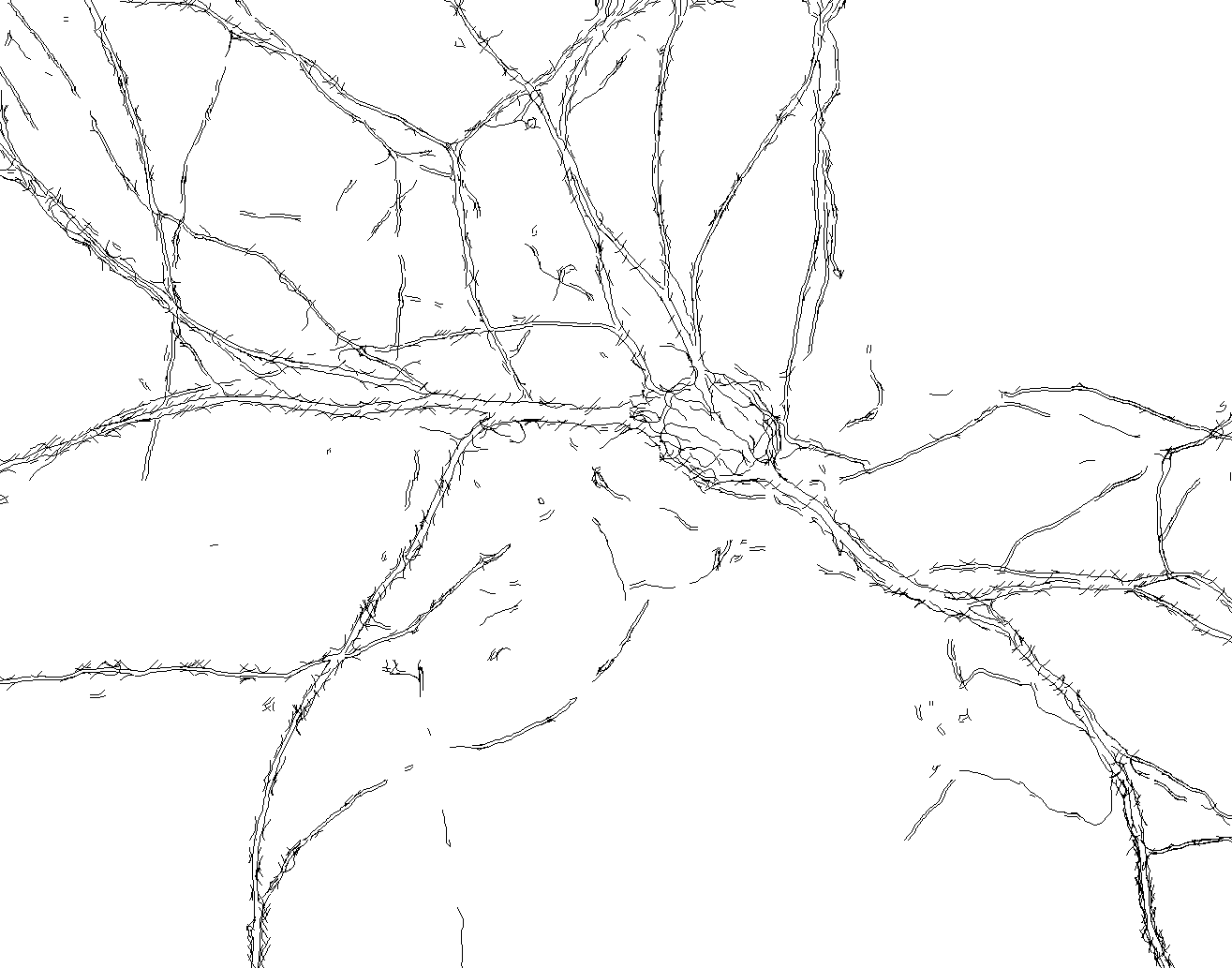}}
\fbox{\includegraphics[width=1.8cm]{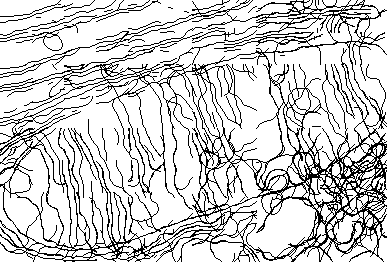}}
\fbox{\includegraphics[width=1.7cm]{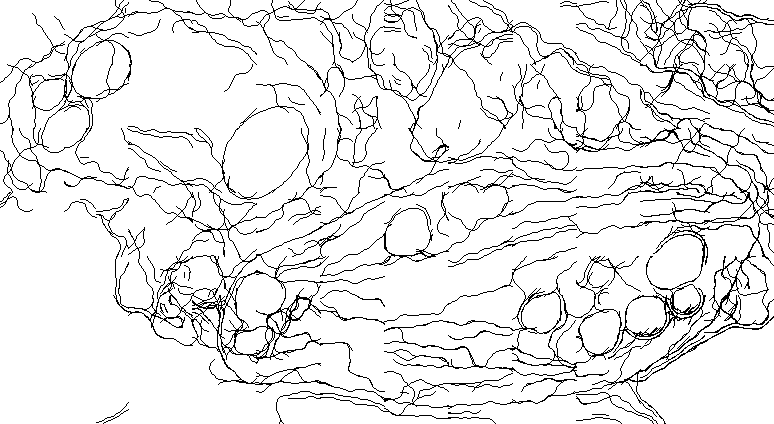}}
\fbox{\includegraphics[width=1.8cm]{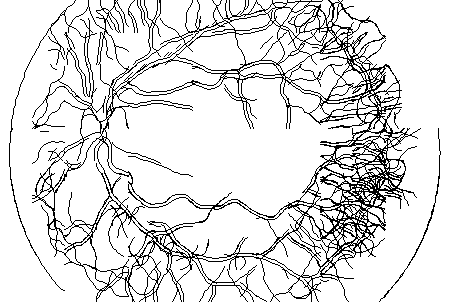}}
\fbox{\includegraphics[width=1.7cm]{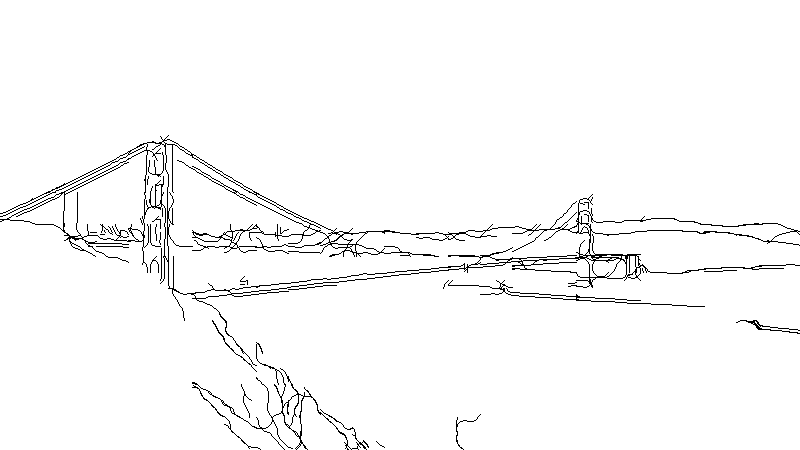}}
\\[0.1cm]
\fbox{\includegraphics[width=1.6cm]{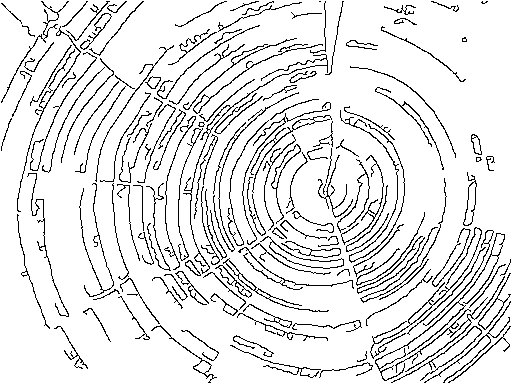}}
\fbox{\includegraphics[width=1.7cm]{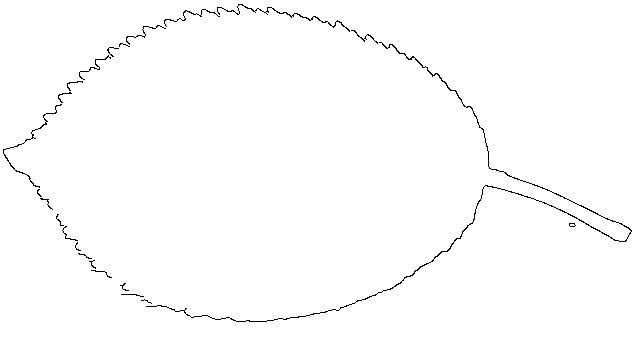}}
\fbox{\includegraphics[width=1.6cm]{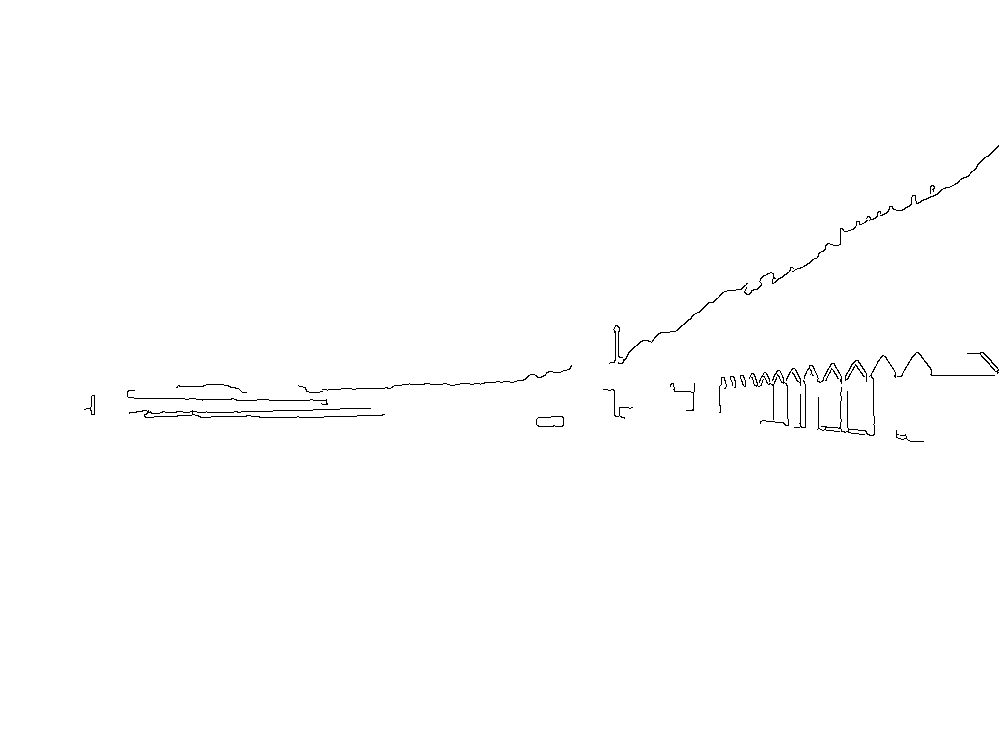}}
\fbox{\includegraphics[width=2cm]{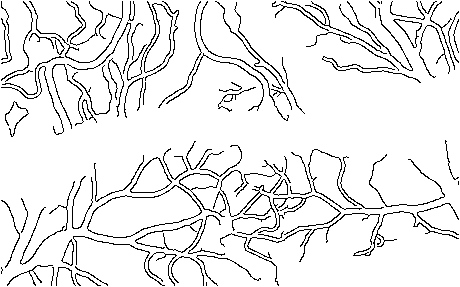}}
\fbox{\includegraphics[width=1.6cm]{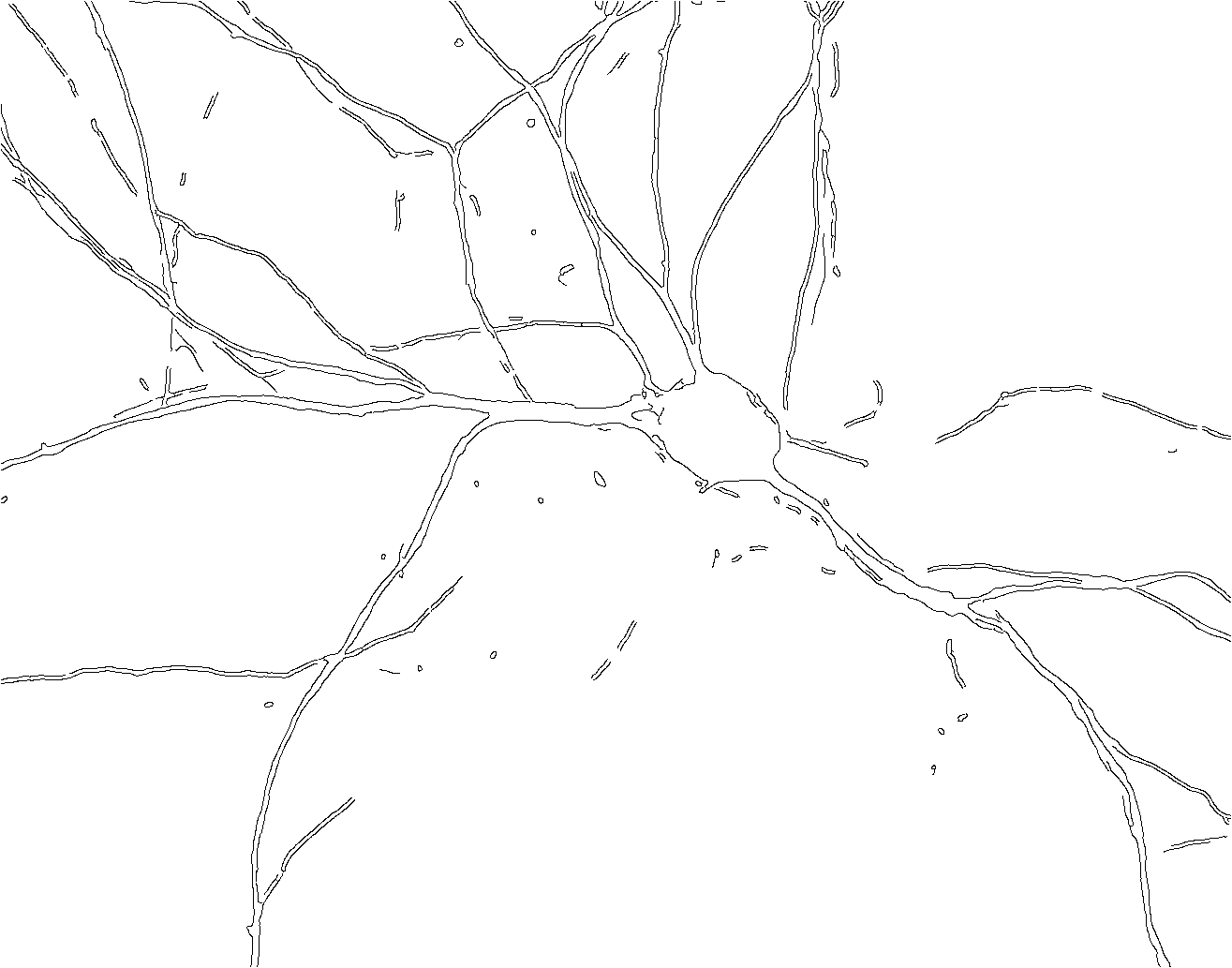}}
\fbox{\includegraphics[width=1.8cm]{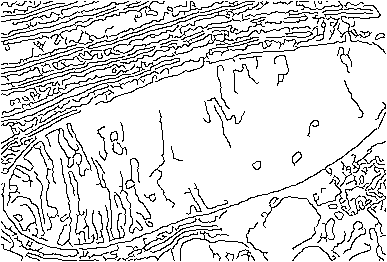}}
\fbox{\includegraphics[width=1.7cm]{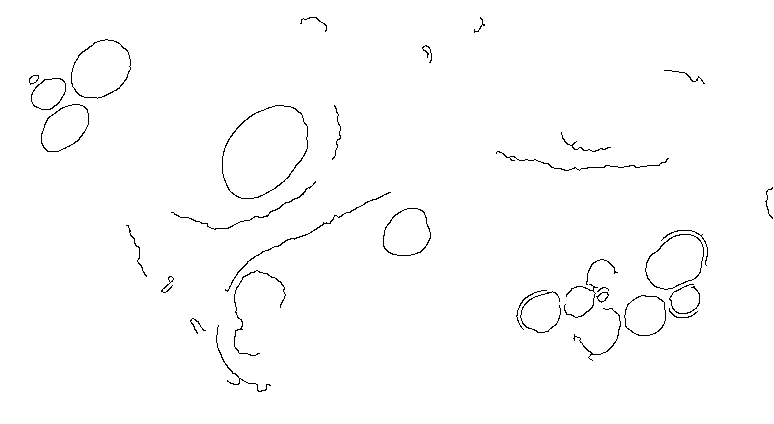}}
\fbox{\includegraphics[width=1.8cm]{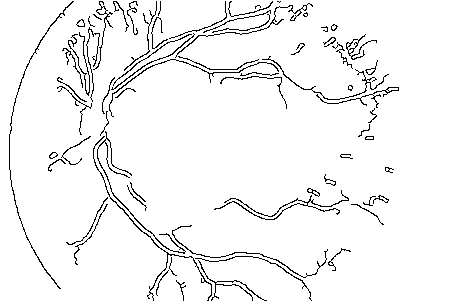}}
\fbox{\includegraphics[width=1.7cm]{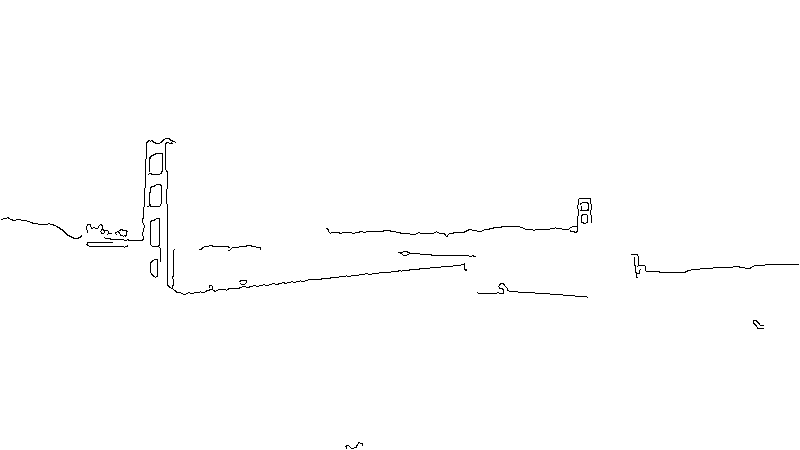}}
\\[0.1cm]
\fbox{\includegraphics[width=1.6cm]{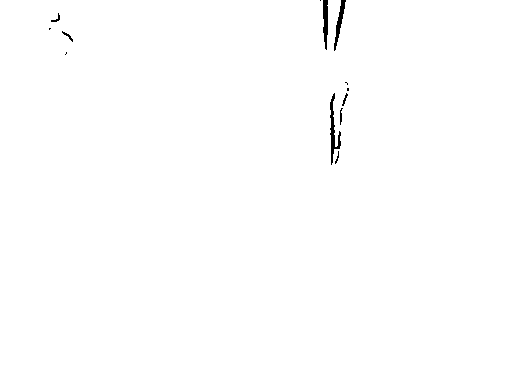}}
\fbox{\includegraphics[width=1.7cm]{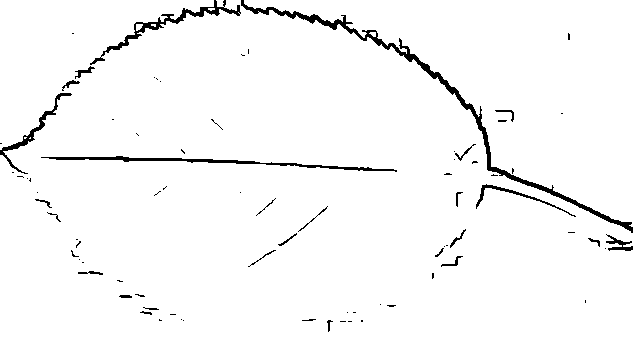}}
\fbox{\includegraphics[width=1.6cm]{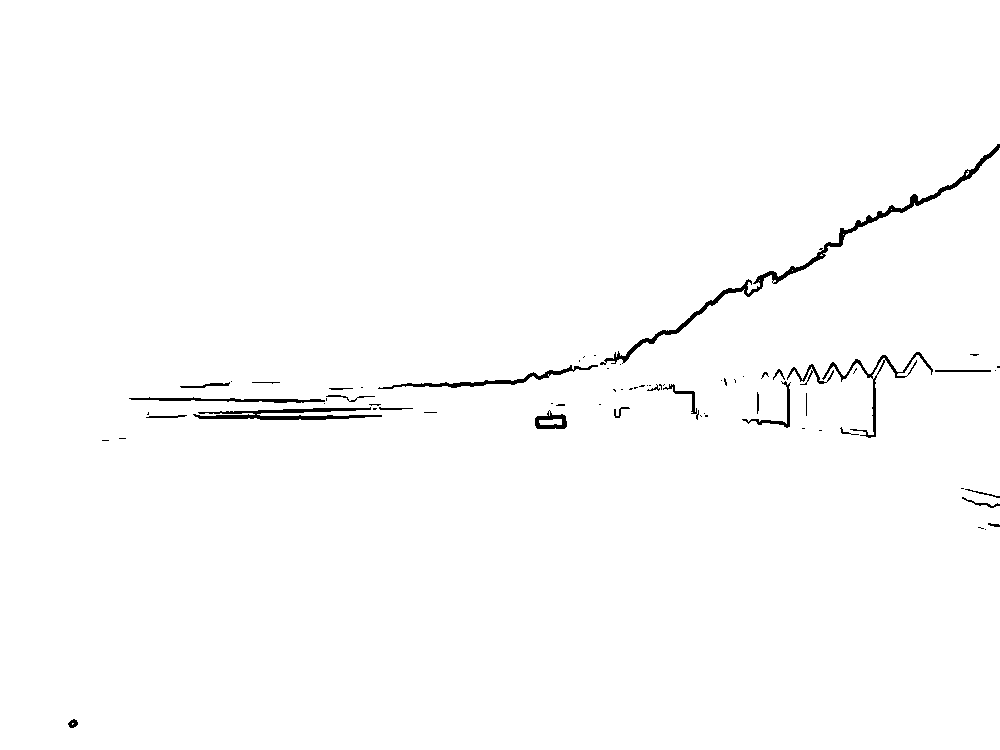}}
\fbox{\includegraphics[width=2cm]{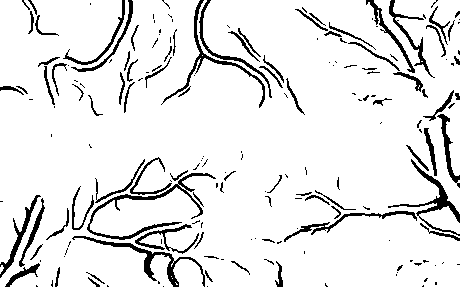}}
\fbox{\includegraphics[width=1.6cm]{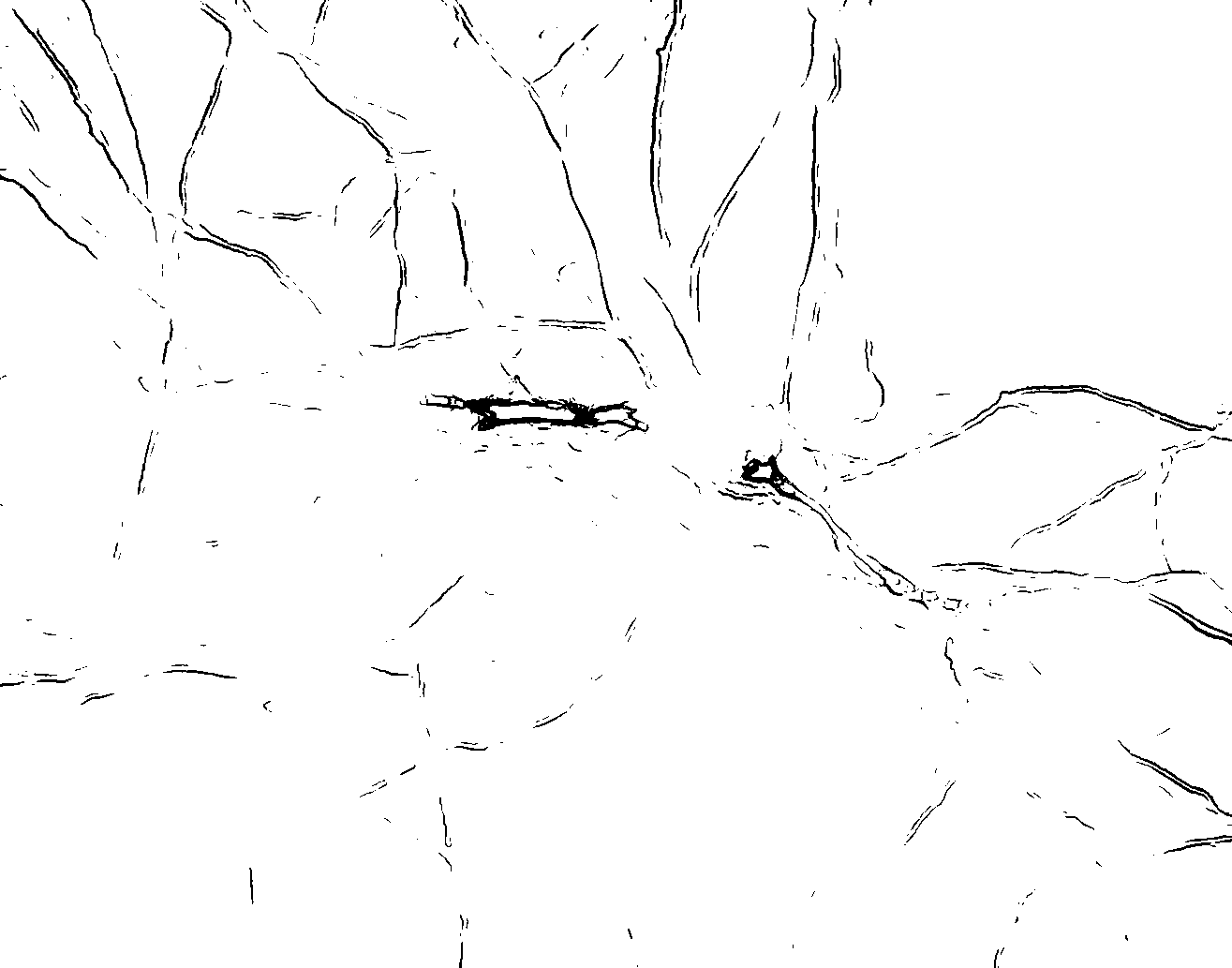}}
\fbox{\includegraphics[width=1.8cm]{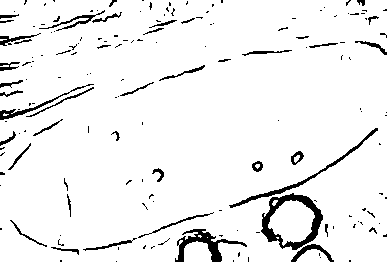}}
\fbox{\includegraphics[width=1.7cm]{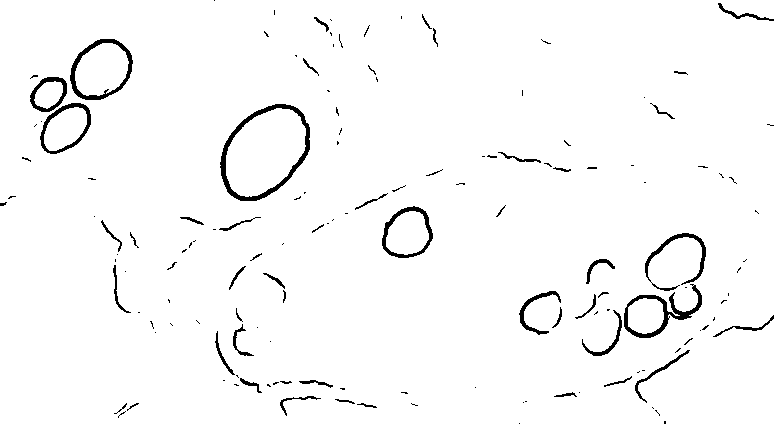}}
\fbox{\includegraphics[width=1.8cm]{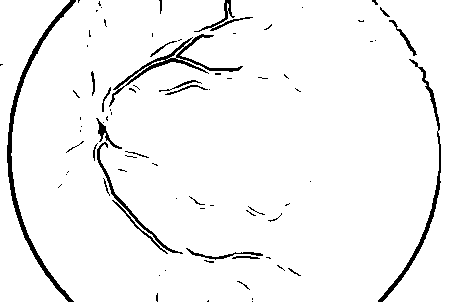}}
\fbox{\includegraphics[width=1.7cm]{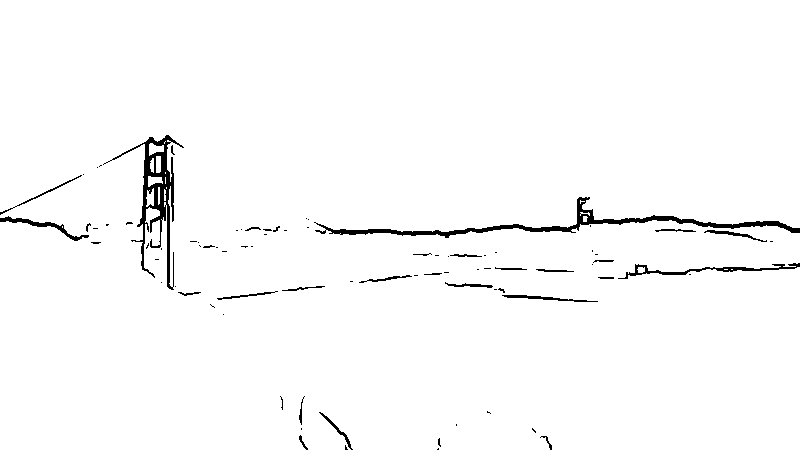}}
\end{center}
\vspace*{-0.25cm}
\caption{From top to bottom, each column shows a real image and the binary results of our $O(N^{1.5})$ Curves method,  Canny and PMI.}
\vspace*{-0.3cm}
\label{fig:realImages}
\end{figure*}

 Fig.~\ref{fig:realImages} shows the results of the various considered  algorithms on several real images taken under relatively poor conditions. It can be seen that our method accurately detect details in these challenging images beyond those detected by existing approaches. Specifically, our method depicts nearly all the growth rings of the tree, the branchings of the nerve cell, and the blood vessels in the retina.

\subsection{Noisy biomedical images}

\begin{figure}
\centering
\fbox{\includegraphics[width=2.5cm]{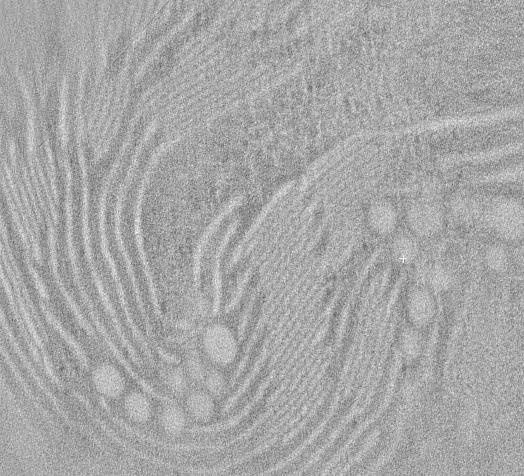}}
\fbox{\includegraphics[width=2.5cm]{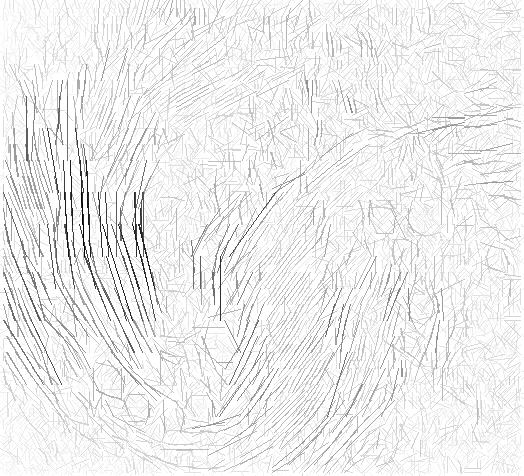}}
\fbox{\includegraphics[width=2.5cm]{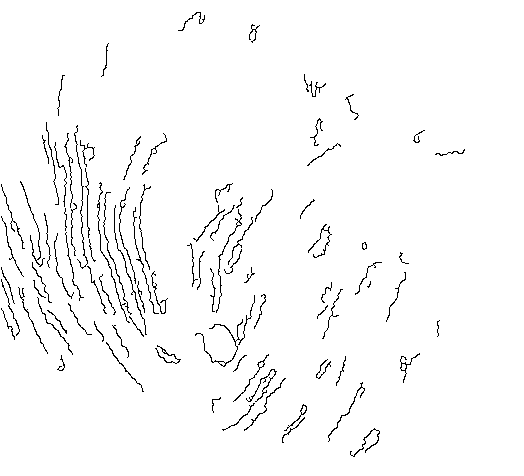}}
\\ [0.1cm]
\fbox{\includegraphics[width=2.5cm]{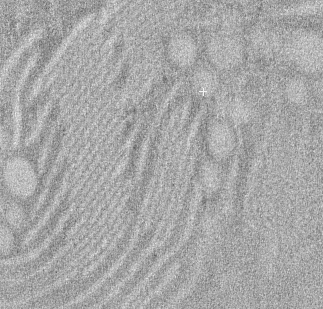}}
\fbox{\includegraphics[width=2.5cm]{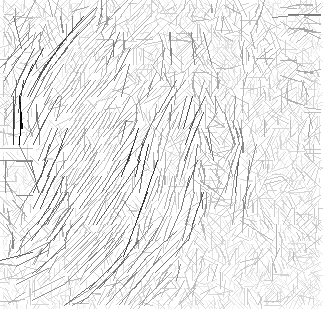}}
\fbox{\includegraphics[width=2.5cm]{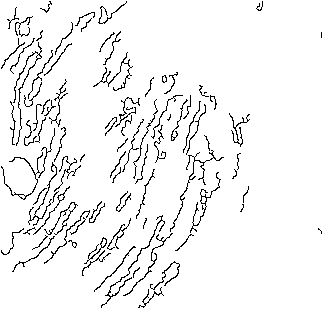}}
\\ [0.1cm]
\fbox{\includegraphics[width=2.5cm]{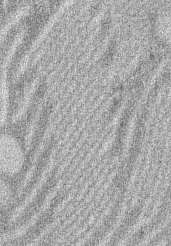}}
\fbox{\includegraphics[width=2.5cm]{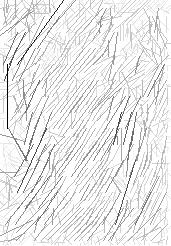}}
\fbox{\includegraphics[width=2.5cm]{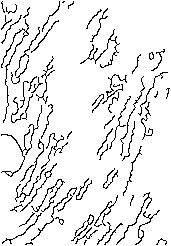}}
\\ [0.1cm]
\fbox{\includegraphics[width=2.5cm]{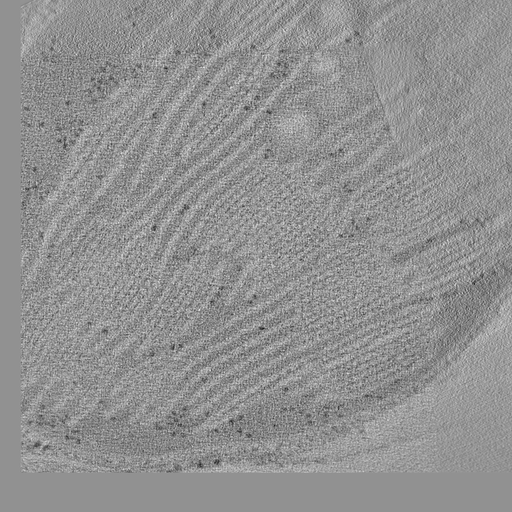}}
\fbox{\includegraphics[width=2.5cm]{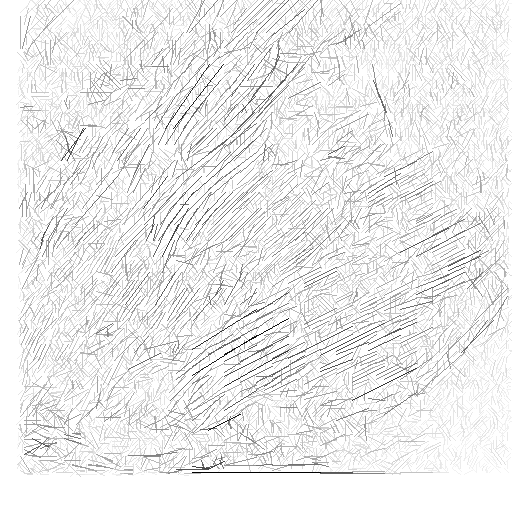}}
\fbox{\includegraphics[width=2.5cm]{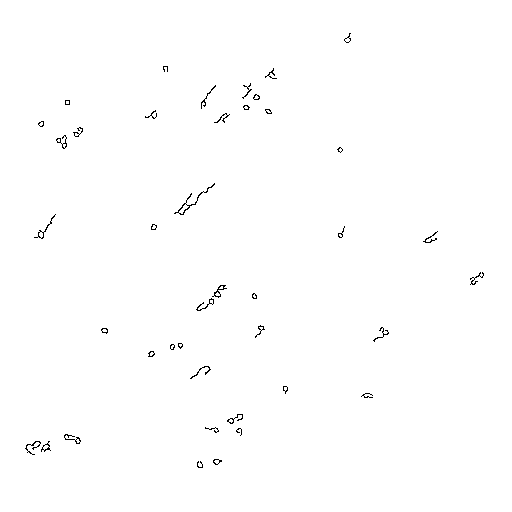}}
\\ [0.1cm]
\caption{Images acquired by electron microscope. From left to right, the original image, the result of our straight line algorithm and  of Canny.}
\label{fig:microscope_results}
\end{figure}

We further tested our algorithms on noisy biomedical images. Fig.~\ref{fig:microscope_results} shows  images of plant cells under certain photosynthetic conditions acquired by an electron microscope. Depending on light intensity, the membranes, which are sub-cell organs (organelles), can get very close to each other. As demonstrated in the figure, our straight edge detector performs well on these images, detecting the edges of the membranes at almost any length and density. In contrast, the Canny edge detector and SE fail to detect the dense edges.

\if false
Finally, although our algorithms are intended to work primarily with non-textured, noisy images, we show in Figure~\ref{fig:BSD} several edge detection results on natural images.


\begin{figure*}
\centering
\fbox{\includegraphics[width=2.8cm]{../imgs/24.png}}~
\fbox{\includegraphics[width=2.8cm]{../imgs/res/res1/24v.png}}~
\fbox{\includegraphics[width=2.8cm]{../imgs/res/res2/24v.png}}~
\fbox{\includegraphics[width=2.8cm]{../imgs/res/res3/24v.png}}~
\fbox{\includegraphics[width=2.8cm]{../imgs/res/res4/24v.png}}~
\\[0.1cm]
\fbox{\includegraphics[width=2.8cm]{../imgs/25.png}}~
\fbox{\includegraphics[width=2.8cm]{../imgs/res/res1/25v.png}}~
\fbox{\includegraphics[width=2.8cm]{../imgs/res/res2/25v.png}}~
\fbox{\includegraphics[width=2.8cm]{../imgs/res/res3/25v.png}}~
\fbox{\includegraphics[width=2.8cm]{../imgs/res/res4/25v.png}}~
\\[0.1cm]
\fbox{\includegraphics[width=2.8cm]{../imgs/31.png}}~
\fbox{\includegraphics[width=2.8cm]{../imgs/res/res1/31v.png}}~
\fbox{\includegraphics[width=2.8cm]{../imgs/res/res2/31v.png}}~
\fbox{\includegraphics[width=2.8cm]{../imgs/res/res3/31v.png}}~
\fbox{\includegraphics[width=2.8cm]{../imgs/res/res4/31v.png}}~
\\[0.1cm]
\fbox{\includegraphics[width=2.8cm]{../imgs/40.png}}~
\fbox{\includegraphics[width=2.8cm]{../imgs/res/res1/40v.png}}~
\fbox{\includegraphics[width=2.8cm]{../imgs/res/res2/40v.png}}~
\fbox{\includegraphics[width=2.8cm]{../imgs/res/res3/40v.png}}~
\fbox{\includegraphics[width=2.8cm]{../imgs/res/res4/40v.png}}~
\\[0.1cm]
\fbox{\includegraphics[width=2.8cm]{../imgs/45.png}}~
\fbox{\includegraphics[width=2.8cm]{../imgs/res/res1/45v.png}}~
\fbox{\includegraphics[width=2.8cm]{../imgs/res/res2/45v.png}}~
\fbox{\includegraphics[width=2.8cm]{../imgs/res/res3/45v.png}}~
\fbox{\includegraphics[width=2.8cm]{../imgs/res/res4/45v.png}}~
\\[0.1cm]
\fbox{\includegraphics[width=2.8cm]{../imgs/78.png}}~
\fbox{\includegraphics[width=2.8cm]{../imgs/res/res1/78v.png}}~
\fbox{\includegraphics[width=2.8cm]{../imgs/res/res2/78v.png}}~
\fbox{\includegraphics[width=2.8cm]{../imgs/res/res3/78v.png}}~
\fbox{\includegraphics[width=2.8cm]{../imgs/res/res4/78v.png}}~
\\[0.1cm]
\fbox{\includegraphics[width=2.8cm]{../imgs/104.png}}~
\fbox{\includegraphics[width=2.8cm]{../imgs/res/res1/104v.png}}~
\fbox{\includegraphics[width=2.8cm]{../imgs/res/res2/104v.png}}~
\fbox{\includegraphics[width=2.8cm]{../imgs/res/res3/104v.png}}~
\fbox{\includegraphics[width=2.8cm]{../imgs/res/res4/104v.png}}~
\\[0.1cm]
\fbox{\includegraphics[width=2.8cm]{../imgs/123.png}}~
\fbox{\includegraphics[width=2.8cm]{../imgs/res/res1/123v.png}}~
\fbox{\includegraphics[width=2.8cm]{../imgs/res/res2/123v.png}}~
\fbox{\includegraphics[width=2.8cm]{../imgs/res/res3/123v.png}}~
\fbox{\includegraphics[width=2.8cm]{../imgs/res/res4/123v.png}}~
\\[0.1cm]

\caption{Real images: Each row shows the original image
and a comparison of our results with those obtained by other algorithms.
From left to right: our curve detection algorithm, our straight line detection algorithm, Canny and SE.}

\label{fig:BSD}
\end{figure*}
\fi

\section{Fiber detection and enhancement}
\label{sec:fiber-enhancement}

We have also extended our method to detect and enhance elongated
fibers. Fibers appear in many types of images, particularly in
biomedical imagery and airborne data. Automatic detection of fibers can save intensive manual labor, e.g., in biomedical research. Detection of fibers can be
difficult due to both imaging conditions (noise, illumination
gradients) and shape variations, such as changes in widths and branchings.
We propose a method that uses edges, 
along with their sign (transition from dark to light or vice
versa) to highlight the fibers.

Specifically, we begin by detecting all edges in the image using our straight edge detector. We next mark those edges as either ``red'' or ``blue'', by classifying their gradient
orientation relative to a pre-specified canonical orientation. We
then construct two diffusion maps, one for the red and the other
for the blue edge maps, by convolving the binary edge map with a
gaussian filter ($\sigma_x^2 = 2.25$ and $\sigma_y^2 = 1$).
Finally, we multiply the two diffusion maps to obtain the enhanced fibers. This process is analogous to stochastic completion of contours~\cite{Mumford} with each edge
point emitting particles to its surroundings. The product of the
diffusion maps reflects the probability of beginning in one edge
and ending in a neighboring edge of opposite gradient sign. This
process is quite robust to varying fiber widths and to fiber
branchings, and although many red and blue responses may appear in
the image, only those that can be paired are enhanced.

\begin{figure}
\centering
\includegraphics[width=2.0cm]{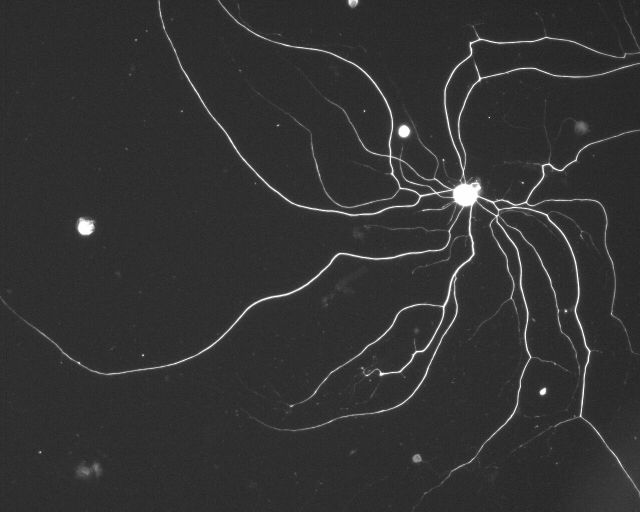}
\includegraphics[width=2.0cm]{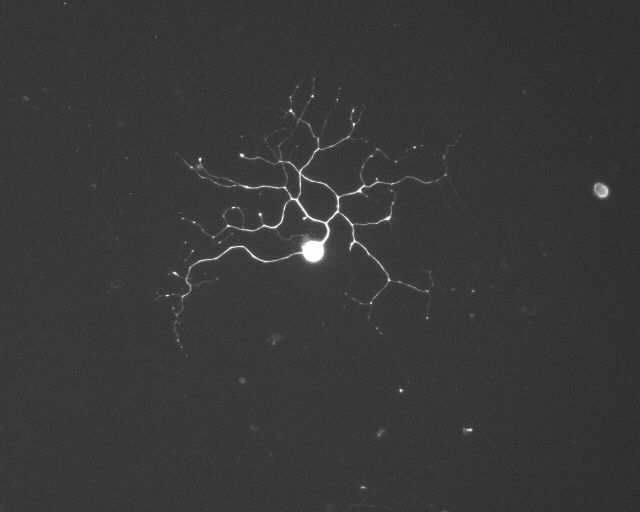}
\includegraphics[width=2.0cm]{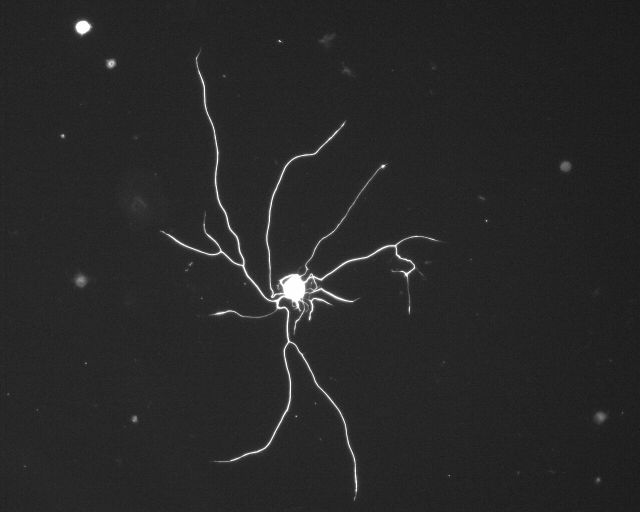}
\includegraphics[width=2.0cm]{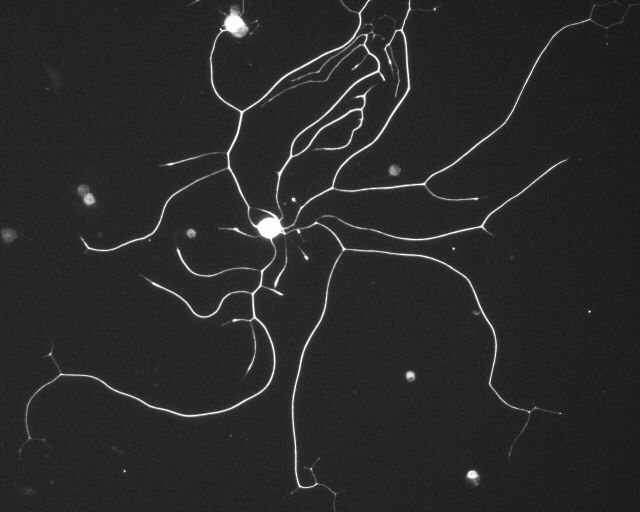}
\\[-1mm]
\includegraphics[width=2.0cm]{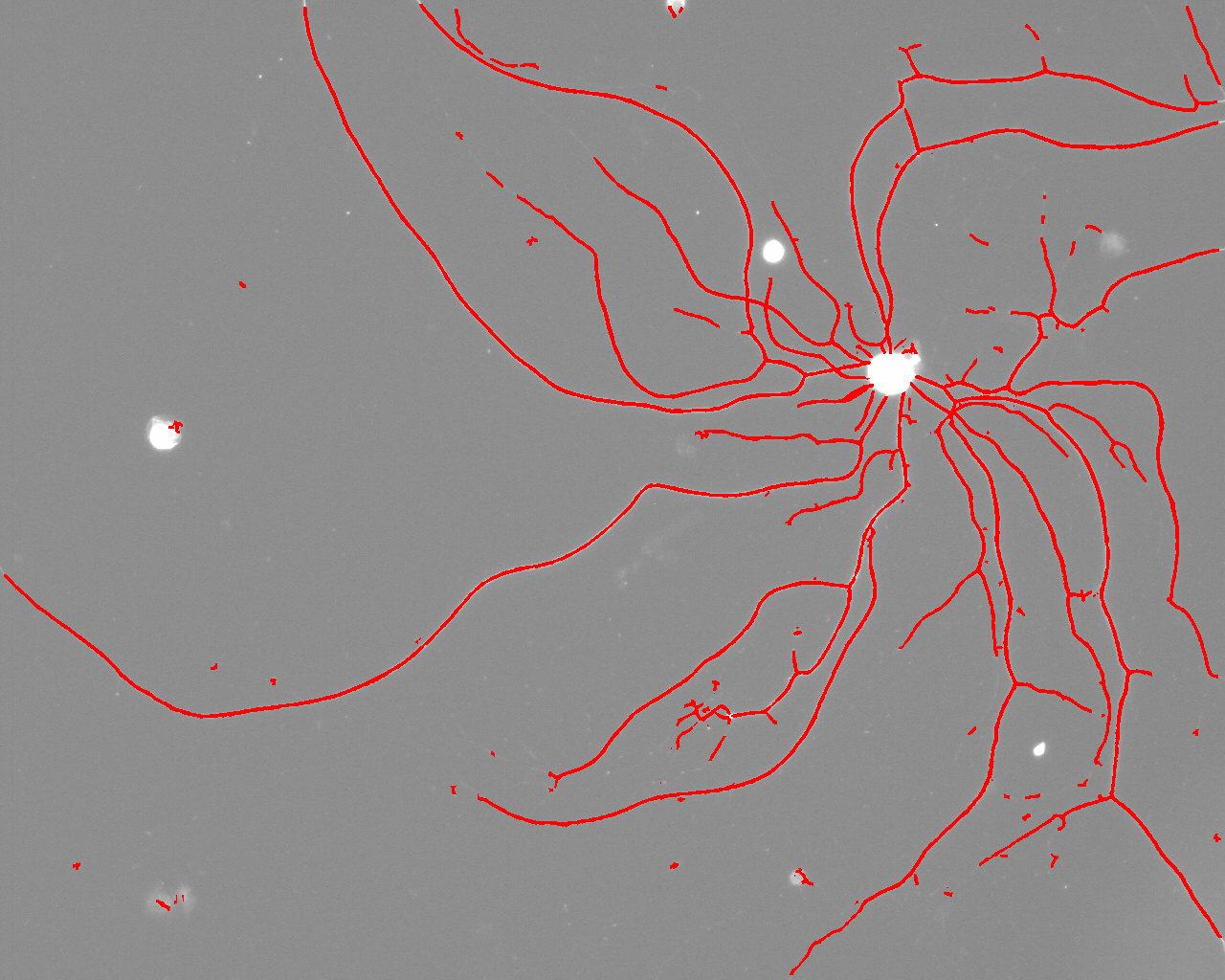}
\includegraphics[width=2.0cm]{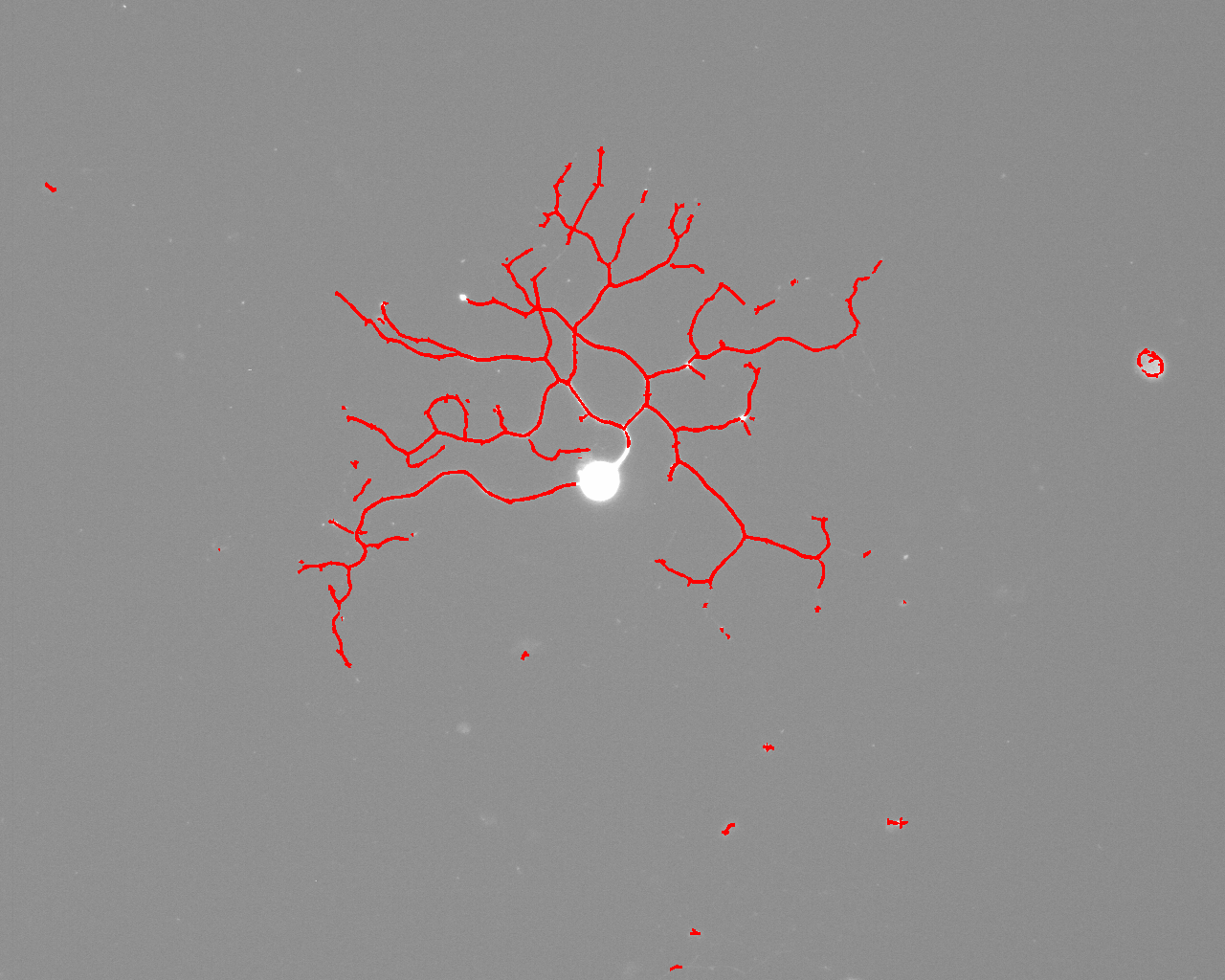}
\includegraphics[width=2.0cm]{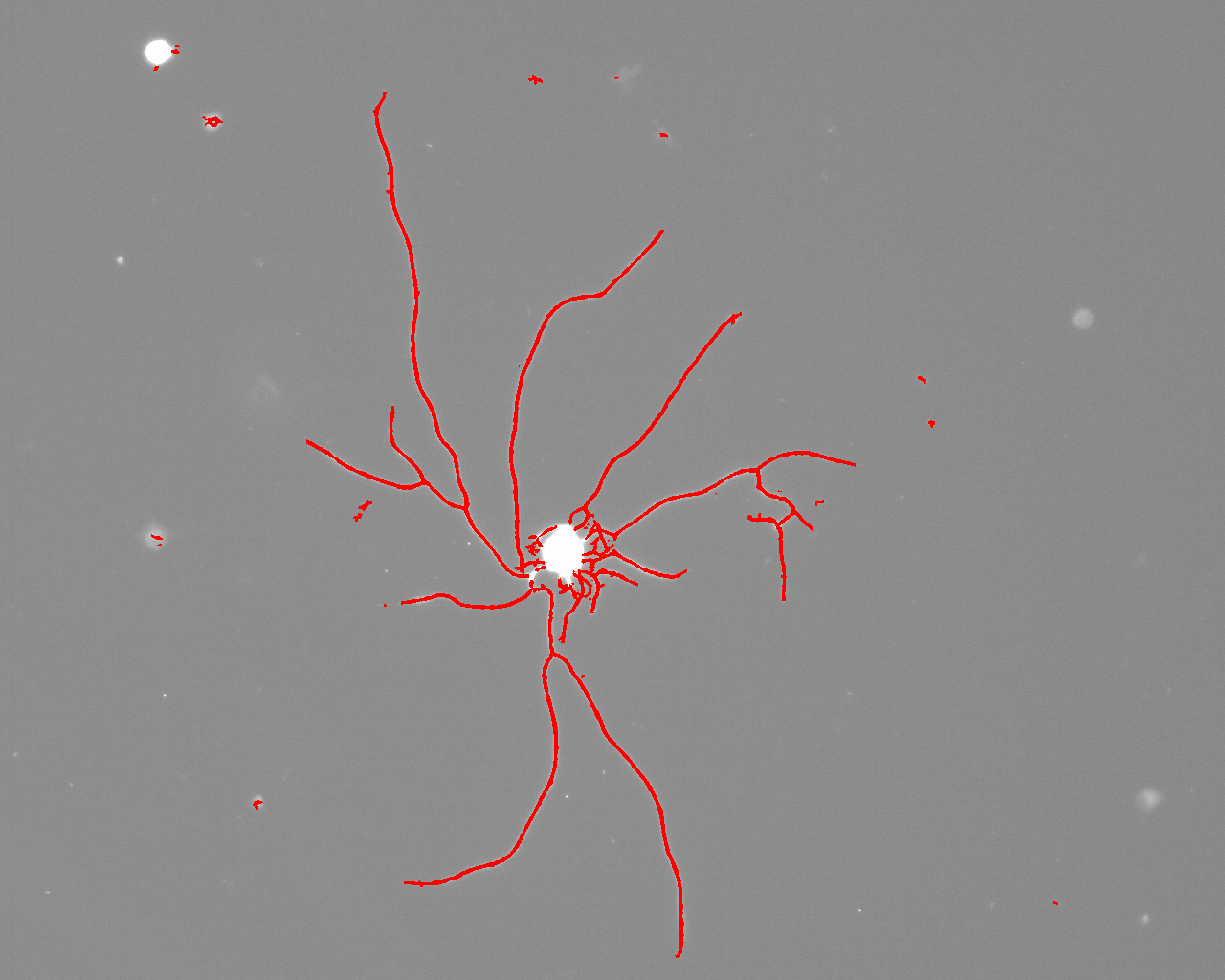}
\includegraphics[width=2.0cm]{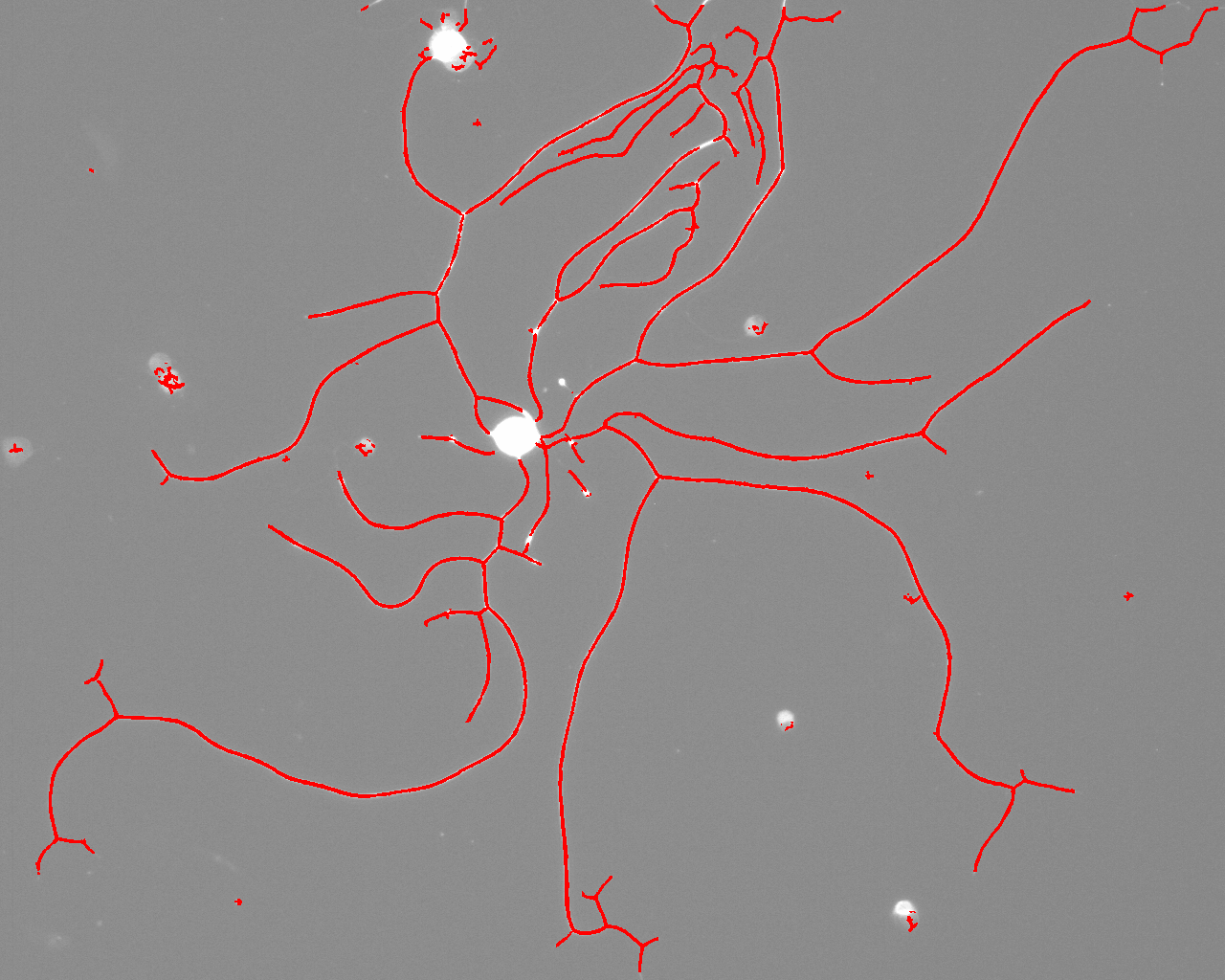}
\caption{Images of nerve cells acquired by light microscopy. Top: original images. Bottom: detected nerve axons overlaid on the original image}
\label{fig:FiberDetection}
\end{figure}

We  applied our fiber enhancement process to fluorescent images of nerve cells acquired by light microscopy\footnote{We thank  Ida Rishal and Michael Fainzilber for the
images.} and used the method to identify branching axonal structures and to measure their total lengths. Comparison of nerve cells under different conditions requires quantitative measurements of cellular morphology, and a commonly used measure is the total axonal length. A single experiment typically produces hundreds of images, many of which suffer from low signal to noise ratio. There is thus an obvious need for fast and accurate automatic processing of such images.
Fig.~\ref{fig:FiberDetection} shows nerve cells images and their enhanced axonal structures. The total axonal length was further computed automatically from the enhanced images and compared to manual length estimation. The two measures match with only slight deviation, see Table~\ref{Table:ManualvsAutomatic}. A more detailed description of this application is provided in~\cite{Ida}.


\begin{table}[ht]
\caption{Measuring total axonal length: manual vs. automatic estimation (in pixel units), for the images in Fig.~\ref{fig:FiberDetection}.}
\begin{center}
\begin{tabular}{|l|c|c|c|c|}
 \hline
{ Manual estimation} & {11940} &
{ 4467} & { 3407} & {7347} \\
\hline
{Automatic estimation} & {11540} & {4295}
& {3844} & {9054} \\
\hline {Relative error (percents)} & {-3.35}
& {-3.85}  & {12.83}  & {23.23}  \\
\hline
\end{tabular}
\label{Table:ManualvsAutomatic}
\end{center}
\end{table}

\section{Conclusion}

We studied the problem of edge detection in noisy images viewing it as search in a space of feasible curves. We showed that the combinatorics of the search space plays a crucial role in the detection of faint edges and subsequently developed two edge detection algorithms, for straight and curved edges. In quest of even faster runtimes on very large and noisy images containing long edges, more recently we extended these methods to have sub-linear runtime. This is achieved by initially processing only small parts of the input image
\cite{horev2015detection,wang2016detecting}.

In future work we hope to further investigate useful shape priors for edges and incorporate those into our formalism. A further challenging problem is to detect texture boundaries in noisy images.

\comment{
\section{Boaz's proof}

{\bf Lemma:} {\em There exists a monotone curve $\Gamma=\Gamma(I)$ of length $L$ starting
at $p_0$, such that
\begin{equation}
\mathbb{E}_I[R(\Gamma)]=\frac\sigma 4\sqrt{\frac2\pi}>0
        \label{eq:Lower_Bound}
\end{equation}
and such that its variance is $O(1/L)$. }

\todo{More details about the variance. Add a drawing. Consider moving to appendix.}

{\em Proof:} To derive Eq. (\ref{eq:Lower_Bound}) we consider a greedy approach to selecting the monotone curve of highest response emanating from a given point $p_0$. Start arbitrarily with flag=up. At each grid point of the curve denote its nearby 3 upper-right pixel values by \(a,b,c\). The next grid point of the curve segment is selected according to the following procedure:

\begin{verbatim}
if flag=up
     if a>0 go up,  flag = up
     else go right, flag = right
else %here flag = right
     if c<0 go right, flag = right
     else go up, flag = up.
end
\end{verbatim}
Recall that by definition the edge response of the resulting curve is the average of its $L$ individual edge filters. Now on approximately half of these, the condition $a>0$ or $c<0$ was satisfied.
Clearly, at each of these segments, the corresponding filter response is strictly positive, and given by $\tfrac12\mathbb{E}[x\,|x>0] $ where $x\sim\mathcal N(0,\sigma^2)$, namely the expectation is equal to
 $\frac\sigma{\sqrt{2\pi}}$.
In contrast, the remaining $\approx L/2$ filter responses, by construction, have zero mean.
Hence, by definition,
$$\mathbb{E}_I[R(\Gamma)] = \frac1L\sum_i \mathbb{E}_I[R(p_i)]=\frac\sigma{4}
\sqrt{\frac2\pi}
$$
As for the variance, note that in roughly $L/2$ of the cases, new pixel values enter consecutive filter responses, rendering them statistically independent. In the remaining approximately $L/2$ segments consisting of corners, the same pixel value enters two consecutive filter responses. Thus, although the exact expression for the variance of this curve is somewhat complicated, it is $O(1/L)$.  \hfill$\Box$

This lemma implies that as $L\to\infty$, $T^\infty$ is indeed
strictly positive. The reason is that using the strong law of large numbers, as $L \rightarrow \infty$ the probability that the response will obtain the value $\frac\sigma4 \sqrt{\frac2\pi}$ approaches 1. This shows that for any reasonable value of $\delta$ (the false detection rate) $T^{\infty}$ must be strictly positive (and $\ge \sigma /4 \sqrt{2/\pi}$). Consequently, faint edges with contrast lower than $T^{\infty}$ cannot be detected unless we allow accepting a considerable number of false positives. Note finally that the main arguments in this section extend also to other (non-gaussian) i.i.d.\ noise models.
}

\subsubsection*{Acknowledgements}

Research was supported in part by the Institute for Future Defense Technologies Research named for the Medvedi, Shwartzman and Gensler Families, and by the European Commission Project IST-2002-506766 Aim Shape. Part of this research was conducted while RB was at TTI-C. At the Weizmann Institute research was conducted at the Moross Laboratory for Vision and Motor Control. We thank Pedro Felzenswalb for sharing his insights with us.

\bibliographystyle{IEEEtran}

\bibliography{bib}

\vspace*{-10mm}

\begin{IEEEbiography}[{\includegraphics[width=1in,height=1.25in,clip,keepaspectratio]{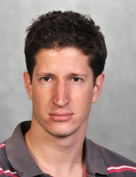}}]
{Nati Ofir} received his BSc in Computer Science at the Hebrew University in 2011 and the MSc degree in mathematics and computer science from the Weizmann Institute of Science in 2013.  He is currently a Ph.D. Candidate at the Weizmann Institute of Science, Israel, in the Department of Computer Science and Applied Mathematics. His research interests are in the areas of computer vision, image processing and machine learning.
\end{IEEEbiography}

\vspace*{-12mm}

\begin{IEEEbiography}[{\includegraphics[width=1in,height=1.25in,clip,keepaspectratio]{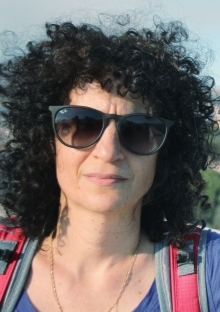}}]
{Meirav Galun} received the M.Sc. and Ph.D. degrees in applied mathematics from the Weizmann Institute of Science in 1992 and 1998, respectively, where she accepted an Excellency Award. She is currently an Associate Staff Scientist in the Applied Mathematics and Computer Science Department at the Weizmann Institute of Science. Her research interests include computer vision, optimization, data analysis, multiscale methods, and biomedical imaging.
\end{IEEEbiography}

\vspace*{-12mm}

\begin{IEEEbiography}[{\includegraphics[width=1in,height=1.25in,clip,keepaspectratio]{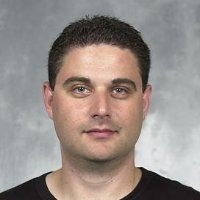}}]
{Sharon Alpert} received the M.Sc. degree summa cum laude in computer science from the Tel-Aviv University in 2003. He received his Ph.D. degree from the Weizmann Institute of Science in 2010 where he continued as postdoc at the Department of Computer Science and Applied Mathematics. His research interests lie in computer vision, specifically in the area of image segmentation, edge detection and object recognition.
\end{IEEEbiography}

\vspace*{-12mm}

\begin{IEEEbiography}[{\includegraphics[width=1in,height=1.25in,clip,keepaspectratio]{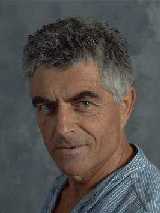}}]
{Achi Brandt} received the M.Sc. degree from the Hebrew University of Jerusalem in 1963 and the Ph.D. from the Weizmann Institute of Science in 1965. He conducted postdoctoral studies at New York University's Courant Institute, and then in 1968 joined the Weizmann Institute of Science where he currently serves as Professor Emeritus. He received the Landau Prize (1978), the Rothschild Prize (1990) and the SIAM/ACM Prize (2005). The citation for this prize reads it was awarded ``for pioneering modern multilevel methods, from multigrid solvers for partial differential equations to multiscale techniques for statistical physics, and for influencing almost every aspect of contemporary computational science and engineering.''
\end{IEEEbiography}

\vspace*{-10mm}

\begin{IEEEbiography}[{\includegraphics[width=1in,height=1.25in,clip,keepaspectratio]{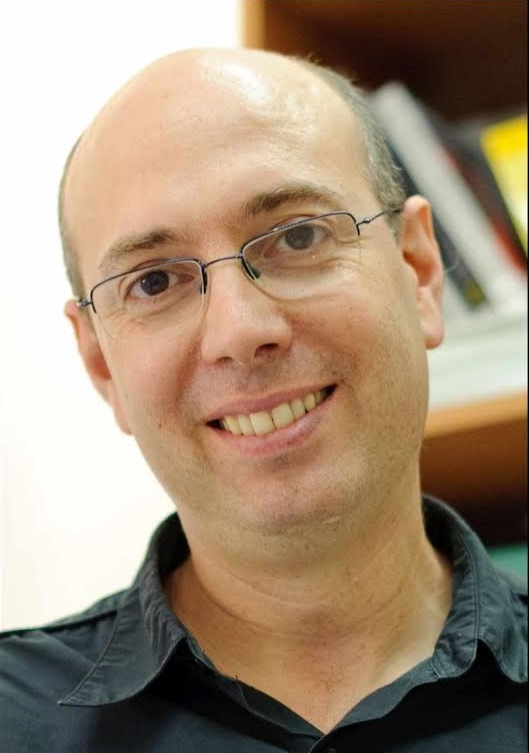}}]
{Boaz Nadler} holds a Ph.D. in applied mathematics from Tel-Aviv University, Israel. After three years as a Gibbs instructor/assistant professor at the mathematics department at Yale, he joined the department of computer science and applied mathematics at the Weizmann Institute of Science, where he is currently associate professor. His research interests are in mathematical statistics, machine learning, and applications in optics, signal and image processing.
\end{IEEEbiography}

\vspace*{-12mm}

\begin{IEEEbiography}[{\includegraphics[width=1in,height=1.25in,clip,keepaspectratio]{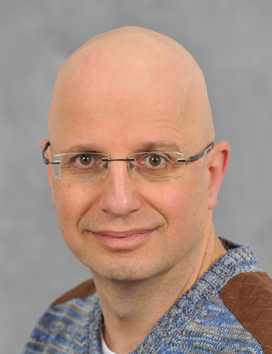}}]
{Ronen Basri} received the Ph.D. degree from the Weizmann Institute of Science in 1991. From 1990 to 1992, he was a postdoctoral fellow at the Massachusetts Institute of Technology. Since then, he has been affiliated with the Weizmann Institute of Science, where he currently holds the position of professor and the Head of the Department of Computer Science and Applied Mathematics. His research has focused on the areas of object recognition, shape reconstruction, lighting analysis, and image segmentation.
\end{IEEEbiography}

\end{document}